\documentclass[logo,msc,ai]{infthesis}           
\usepackage{msccheck}
\usepackage{graphicx}

\usepackage{wrapfig}
\usepackage{hyperref}
\usepackage{subcaption}
\usepackage[nocompress]{cite}

\usepackage{listings}
\usepackage{amsmath}
\usepackage{amssymb}
\usepackage{multirow}
\usepackage{algorithm}
\usepackage{algpseudocode}

\usepackage{microtype} 

\begin{document}
\begin{preliminary}

\title{Enhancing Reinforcement Learning in 3D Environments through Semantic Segmentation: A Case Study in ViZDoom}

\author{Hugo Huang}

\submityear{2024}
\graduationdate{November 21, 2024}

\abstract{
Reinforcement learning (RL) in 3D environments with high-dimensional sensory input poses two major challenges: (1) the high memory consumption induced by memory buffers required to stabilise learning, and (2) the complexity of learning in partially observable Markov Decision Processes (POMDPs). This project addresses these challenges by proposing two novel input representations: \textbf{SS-only} and \textbf{RGB+SS}, both employing semantic segmentation on \textbf{RGB} colour images. Experiments were conducted in deathmatches of ViZDoom \cite{ViZDoom}, utilizing perfect segmentation results for controlled evaluation. Our results showed that \textbf{SS-only} was able to reduce the memory consumption of memory buffers by at least 66.6\%, and up to 98.6\% when a vectorisable lossless compression technique with minimal overhead such as run-length encoding \cite{run-length-encoding} is applied. Meanwhile, \textbf{RGB+SS} significantly enhances RL agents' performance with the additional semantic information provided. Furthermore, we explored density-based heatmapping as a tool to visualise RL agents' movement patterns and evaluate their suitability for data collection. A brief comparison with a previous approach \cite{PreviousSS} highlights how our method overcame common pitfalls in applying semantic segmentation in 3D environments like ViZDoom.
}

\maketitle

\newenvironment{ethics}
   {\begin{frontenv}{Research Ethics Approval}{\LARGE}}
   {\end{frontenv}\newpage}

\begin{ethics}
This project was planned in accordance with the Informatics Research
Ethics policy. It did not involve any aspects that required approval
from the Informatics Research Ethics committee.

\standarddeclaration
\end{ethics}

\begin{acknowledgements}
First and foremost, I would like to express my deepest gratitude to my supervisor, Pavlos Andreadis, for his unwavering support throughout this project. I am also deeply thankful to my parents, who patiently listened to all my progress reports despite not having much background in this field of study.

I would also like to extend my thanks to the developers who have contributed to the open-source projects Stable Baselines3, ViZDoom (and also naturally, ZDoom). These projects have been invaluable over the past few months and I wouldn't have been able to complete this project without them.

Lastly, I would like to acknowledge the original developers of Doom for creating such an iconic video game, which hasn't lost any of its magic over the past 30 years. Doom's E1M1 level and its signature music will always hold a special place in my heart, they sparked my interest in retro-gaming and led me to explore the field of reinforcement learning.

The source code for this project is available on GitHub\footnote{\url{https://github.com/Trenza1ore/SegDoom}}\footnote{Parts of the source code have been taken directly from my undergraduate final year project. \cite{FYP}}, with several pre-trained models stored via Git Large File Storage. Additionally, game-play recordings can be accessed on Google Drive\footnote{\url{https://drive.google.com/drive/folders/1KHQZr7Uls9YiFIPP_bxTmTyBuNYZZo-y}}. Utility software library which implements the compressed replay / rollout buffers mentioned in this paper has been published to PyPI\footnote{\url{https://pypi.org/project/sb3-extra-buffers}} and source code is available on GitHub\footnote{\url{https://github.com/Trenza1ore/sb3-extra-buffers}}.

\end{acknowledgements}

\tableofcontents
\end{preliminary}

\chapter{Introduction}

\section{Motivations}
Reinforcement learning (RL) is a popular area in machine learning that aims to model and solve decision-making problems in environments that are dynamic and stochastic. RL has made a significant impact on both the academic and industrial landscapes in recent years, fostering advancements in various fields spanning from Natural Language Processing \cite{RLHF} to robotics \cite{robotics-rl} and intelligent game-playing agents \cite{game-rl}. However, despite the success, RL faces significant challenges. Particularly in the pure-visual domain, with high-dimensional sensory data as input to the algorithms, the complexity involved in learning a partially observable Markov Decision Process (POMDP) \cite{POMDP} and high memory consumption become critical bottlenecks. This project seeks to address the above-mentioned issues by proposing two novel input representations based on semantic segmentation: one that reduces memory usage of memory buffers by 66.6\% initially but can be further optimised to 98.6\%\footnote{When compressed at a vectorisable O(n) time complexity via run-length encoding \cite{run-length-encoding}.} while maintaining RL agents' performance, and another that improves performance significantly with the presence of additional semantic information. \emph{To our best knowledge, this is the first work to propose an input representation that reduces memory usage at this magnitude for reinforcement learning in 3D environments.}

\subsection{The Memory Consumption Problem}
Most recent advancements in RL can be further categorized as Deep Reinforcement Learning (DRL) as they would utilize deep neural networks (DNN) \cite{cnn-yann-lecun-1998} in the decision-making process. Before DQN \cite{DQNAtari}, DRL was considered unstable and would even diverge when the action-value function (also known as Q function \cite{Q-Learning}) is represented with a nonlinear function approximation such as DNNs \cite{DRL-instability}. Many algorithms proposed to stabilise DRL's learning process utilized a technique known as experience replay \cite{ExperienceReplay}. Experience replay would store the data from n-latest time-steps in a replay buffer and during training, the RL algorithms sample data from this buffer randomly, either following a uniform distribution of probability or a biased one with various definitions of priorities to improve sample efficiency \cite{prioritized-experience-replay}. 

Experience replay helps to stabilise learning by removing the correlations that are presented in the sequence of observations acting as input to the RL agents \cite{DQNAtari}, reducing the probability of the agents overfitting to recent experiences. This benefit comes at the cost of memory consumption when input data is high-dimensional such as image or video. Policy gradient \cite{policy-gradient} RL algorithms such as trust region policy optimization (TRPO) \cite{trpo} or proximal policy optimization (PPO) \cite{PPO} typically do not make use of experience replay, however, they may still make use of similar memory buffer components. These memory buffers share the same memory consumption concerns as the replay buffer despite serving different purposes.

\subsection{Importance of Understanding Visual Input}
In RL, especially with environments closely mimicking real-world scenarios, the ability of an agent to interpret visual information is crucial for high-quality decision-making. Visual perception is crucial to transferring the knowledge of RL agents that have been trained in virtual, computer-simulated environments to real-world applications in areas such as autonomous driving and robotics. This is where semantic segmentation comes into play, as it provides additional information to label and categorise different elements in the current scene, much like how we humans would naturally do by instinct.

As a human, we can understand objects in the environment with different appearances than what was previously known to us, this stems from our ability to assign labels to these unseen entities describing what class of objects they belong to \emph{semantically}, with relation to other objects presented in the scene. For example, when one is playing a video game and sees a new demonic figure appearing on-screen while holding an object that looks like it can be used as a "weapon", we naturally correlate this information of us holding a weapon to the demonic appearance and would draw the conclusion that "this is a new enemy" and potentially also a slight hint of "I should try to attack it". Semantic segmentation (SS) is a task in machine learning that models this behaviour/ability, taking images as input and for every pixel in the input, the SS model would output the confidence level for each semantic class this pixel might belong to. For each pixel, the class with the highest confidence level is selected to produce an SS mask, which labels the semantic class of every pixel in an image.

\subsection{Our Proposed Methods}
To address the challenges of high memory consumption and improve performance in visually complex/noisy 3D environments, this project proposes two new representations of the input observation for RL agents. By using SS masks either as a replacement or an augmentation to the well-adopted \textbf{RGB} colour images in previous literature \cite{ViZDoom, DRQNDoom}, we aim to significantly reduce memory usage with the \textbf{SS-only} representation and enhance the agents' robustness and decision-making capabilities with \textbf{SS+RGB}:
\begin{enumerate}
    \item \textbf{SS-only}: a predicted SS mask is used as input to RL agents directly, reducing memory consumption of memory buffers by cutting the number of colour channels to 1/3 with reduced bit-depth for each pixel. This also adds new compression potentials for techniques such as run-length encoding (RLE) \cite{run-length-encoding} which can further reduce memory consumption to less than 2\% of \textbf{RGB} (see table \ref{tab:mem-cmp}).
    \item \textbf{SS+RGB}: a predicted SS mask is added as the fourth colour channel to augment the \textbf{RGB} image and provide semantic information to the RL agents.
\end{enumerate}

We chose to evaluate the effectiveness of our proposed method in Doom, a 3D First-Person-Shooting (FPS) video game commonly used as benchmark for visual-perception-based control in 3D environments \cite{ViZDoom, DRQNDoom}. The ViZDoom \cite{ViZDoom} platform is an open-source project built upon ZDoom \cite{zdoom}, a source port of the original Doom game engine. ViZDoom is designed as a tool for visual-perception-based machine learning and provides API for direct access to the game engine, allowing perfect, ground-truth SS results to be obtained.

In this project, all of our experiments are run with the PPO \cite{PPO} model as our RL agent, as it was well-known for being stable \cite{DRL-that-matters} and has been shown to work well with ViZDoom scenarios in literature\cite{PreviousSS}. A DeepLabV3 model \cite{rethinkatrous} with ResNet-101 \cite{resnet} backbone was trained to perform the semantic segmentation task. To evaluate its real-world performance, we adopted the mean intersection over union (mIoU) metric. According to our measurement, the mIoU of its predictions with ground truth on each map ranged from 0.846\footnote{In the trained map (Map 1).} to 0.683\footnote{In a complex, unseen map with structures that were not present in training data (Map 3).} in actual game-play sessions of the SS+RGB agent.

Our results showed that the SS+RGB variants outperformed the baseline RGB agents significantly in all evaluated scenarios, both seen and unseen. A previous state-of-the-art paper \cite{DRQNDoom} stated that using RGB colour images (3 colour channels) as input yielded better performance than grayscale images (1 colour channel) in ViZDoom, our proposed SS-only representation also uses one channel only, yet yielded comparable performance with baseline RGB models while cutting down memory consumption significantly. Both the proposed and baseline methods surpassed the best results of the built-in ZCajun bots\footnote{This bot was not well-document due to its age, information of it can still be found at this website (as of the writing of this dissertation in August 2024): \href{https://www.doomworld.com/mellow/bots.shtml}{https://www.doomworld.com/mellow/bots.shtml}}, which used node-based navigation and had near-perfect information about the whole map. \emph{Note that both the SS model and RL agents have been trained on only one map\footnote{We chose Map 1, a map used by a similar study \cite{PreviousSS}.} to better represent real-world situations where unseen data is common.}

\section{Related Works \& Novelty of Our Proposed Methods}
A simple example of SS improving visual-perception-based control performance in RL is 2D video game playing, a previous study \cite{SSMario} has shown that utilizing semantic segmentation masks as input to RL agents, their ability to play the Super Mario Bros video game saw a significant improvement in robustness when controlling in unseen environments with similar appearances to the ones RL agents have been trained on and the generalisation performance improved as the SS-augmented agents were able to learn multiple levels simultaneously during training compared to baseline models that attempt to overfit to specific levels and are unable to learn a policy that consistently performs well in multiple levels.

For a more complex example in the field of robotics, research has shown that SS enables RL agents to transfer a learnt control policy from an indoor environment to an outdoor environment during evaluation effectively and outperformed every RL agent that used RGB images or depth maps as input \cite{SSRobotics}.

While a previous study \cite{PreviousSS} has explored replacing RGB inputs with an SS-based representation in ViZDoom deathmatches using PPO-based RL agents, our work made several novel contributions and addressed some potential issues in the previous work that led to its unexpected results, which showed only a marginal improvement for using DNN-predicted SS compared to normal RGB colour input. We attribute the lower-than-expected improvement to multiple potential factors instead of the unsuitability of semantic segmentation in 3D environments:
\begin{enumerate}
    \item They used an SS model (a DeepLabV3+ model \cite{DeepLabV3+}, ResNet-101 as backbone \cite{resnet}) with a higher resolution for input images than their PPO agents, the additional down-sampling (the predicted SS masks is already up-sampled twice within DeepLabV3+) may induce unwanted artefacts and loss of critical information.
    \item An arbitrary colour palette was used to map the predicted SS masks back to RGB colour space, the colour palette may not be ideal in representing relationships between semantic classes as neither the Euclidean distance nor the cosine similarity between RGB values of different classes correspond to a meaningful measurement of their actual semantic similarities. 
    \item The previous issue is further amplified by the RGB-based down-sampling to produce the input image for the RL agents, as common down-sampling algorithms for RGB images typically involve averaging or interpolation of pixel values, the colour palette did not account for down-sampling and added additional noise to the RL agents' input.
    \item The training and testing dataset for semantic segmentation may not represent the input game frames in actual game-play sessions and the model may have overfit to the dataset. The reported mIoU was as high as 0.982 yet the performance difference between predicted segmentation and perfect segmentation is more significant than its improvement to the RGB baseline.
\end{enumerate}

\label{avoid-down-sampling}
Despite using only the same map as the previous study to train both the RL agents and the SS model, we avoided the pitfalls of unnecessary down-sampling and arbitrary colour mapping by directly utilizing SS masks in their original form, ensuring the preservation of semantic information while not adding any presumed relationship between semantic classes. Additionally, we evaluated the agents on different unseen maps instead of only the one they were trained on and included an additional frame-stacking option for SS-only agents, utilising the reduced memory usage to our advantage. We also performed an analysis of each agent's behaviour (movement patterns, weapon usage, etc.) to provide insights into how learnt policies are affected by different input representations. Finally, we noticed the compression potential of SS masks and provided evidence for a 98\% reduction in memory usage compared to storing raw RGB input while introducing minimal overhead in table \ref{tab:mem-cmp}.

For semantic segmentation, we opted to use a slightly simpler DeepLabV3 \cite{rethinkatrous} model with the same ResNet-101 backbone for faster inference without the additional decoder structure in DeepLabV3+. To obtain a dataset that is more representative of actual game-play sessions, we performed a heatmap analysis of our initial batch of RL agents trained with perfect SS masks obtained from ViZDoom and cherry-picked an agent that visits every corner of the map most uniformly to build the dataset. Evaluation of our SS model was performed on game-play sessions with our best SS+RGB PPO agent and further analysed with per-class IoU to provide additional insight.

\section{Objectives}
This project is guided by the following hypotheses:
\begin{itemize}
    \item Replacing the three-channel RGB input to RL agents in 3D environments with a one-channel semantic segmentation mask will significantly improve training memory efficiency without substantially reducing performance.
    \item The semantic segmentation mask can also be efficiently and effectively lossless compressed utilising a vectorisable algorithm at O(n) time complexity.
    \item Stacking multiple subsequent SS masks will provide additional temporal information to an RL agent, leading to improved decision-making.
    \item Augmenting the RGB input with an additional semantic segmentation channel will improve RL agents' robustness and performance in 3D environments.
\end{itemize}

Experiments conducted for this project all involve the augmentation or replacement of RGB colour input to RL agents with DNN-predicted SS masks, we would not be converting the predicted 1-channel SS masks back to 3-channel RGB images as they are simply another representation of the SS masks, with additional redundant or meaningless information which may cause the training to be more unstable. 

\chapter{Background}

In this section, we will review the background materials on artificial neural networks, semantic segmentation, reinforcement learning, and applications of AI with video games.

\section{Artificial Neural Networks}
Artificial neural network (ANN) \cite{artificial-neural-network} is a class of machine learning models that draw inspiration from the structure of the human nervous system, especially the neural networks in the brain. A typical ANN consists of three types of layers: \textbf{input layer} that receives raw or pre-processed data as input to the ANN, \textbf{output layer} that produces the final predictions given the input data, and \textbf{hidden layer} that is placed between input and output layers to process and extract high-level features from the input data. 

A typical layer in an ANN contains one or multiple neurons, which act as the smallest unit in an ANN. The neurons of one layer would connect to neurons of subsequent layers, with a weight assigned to each connection. A neuron's "activation value" is calculated as the weighted sum of all incoming connections from previous layers. Thus, the transformations between layers are essentially linear. This type of layer is known as \textbf{fully connected layers} or dense layers.

\subsection{Backpropagation}
The weighted connections between two fully connected layers can be represented as weight matrices and these matrices are updated during the training phase of their ANN models. The most common method for training ANNs is \textbf{backpropagation} \cite{cnn-yann-lecun-1998}, a technique that updates every weight value with a small fraction of its gradient $w.r.t.$ the errors measured between expected output (also known as ground truth) and output of the ANN, the gradient is inversed to minimise error instead of increasing it. The small fraction is kept consistent throughout the whole ANN and is known as the \textbf{learning rate} of the model, controlling the magnitude of gradient updates. 

The name "backpropagation" comes from the back-to-front nature of its operation, the last layer would have its gradients $w.r.t.$ output errors calculated first, and each layer's gradient values are calculated $w.r.t.$ gradient values of the layer behind it, propagating backwards toward the first layer.

\subsection{Deep Neural Networks}
ANNs with multiple hidden layers are also known as \textbf{deep neural networks (DNN)}, it has been shown that ANNs with at least one hidden layer can act as universal approximators for functions given that at least one hidden layer would contain a type of non-linear activation function \cite{nn-as-universal-approximator}.

\subsubsection{Activation Function}
Because multiple linear transformations can be combined into one, there is no benefit in building an ANN with hidden layers and the only functions this type of ANN can approximate are linear transformations. To approximate non-linear transformations, an \textbf{activation function} is typically applied to neurons in non-input layers to break linearity. An activation function is simply a non-linear function that is applied to the activation value of neurons, the most commonly used functions are Sigmoid (logistic sigmoid), Tanh (hyperbolic tangent), and ReLU (rectified linear unit). According to \cite{activation-functions}, the choice of activation functions can have a significant impact on an ANN's ability to converge to an optimum approximator for the specific task.

\subsection{Convolutional Neural Networks}
Convolutional neural networks (CNN) \cite{Neocognitron, cnn-yann-lecun-1998} are a specific type of ANNs that utilise convolutional layers for feature extraction. A convolutional layer replaces the weighted connections (as seen in fully connected layers) with a set of n-dimensional tensors (which are matrices in 2D and vectors in 1D) known as convolution kernels. Each kernel would perform a mathematical operation known as "convolution" with the input to the layer, the convolutions of input data and every kernel are stacked together to form the final output.

Traditionally, kernels with fixed values such as the Sobel operator \cite{sobel-kernel-talk} have been widely used in computer vision tasks like edge detection. These kernels would contain presumed knowledge from human experts and are used to extract features designed by their creators, this process is called "handcrafted feature extraction". The kernels in CNN are automated and learnable instead, they are updated during every backpropagation with other learnable parameters such as weight matrices to capture common patterns and extract high-level features that are considered valuable by the model.

\section{Semantic Segmentation}
Image segmentation is a fundamental task in computer vision that performs the partitioning of a whole image into multiple segments, with each of these segments corresponding to different objects or different parts of an object. Traditionally, non-neural\footnote{Not involving an artificial neural network.} image segmentation techniques focused on exploiting patterns in low-level features like colour and intensity, these approaches are highly dependent on knowledge from human experts and typically perform poorly in unseen data. While some of these techniques such as pixel value thresholding \cite{thresholding} or kernel-based edge detection \cite{edge-detection-via-kernel-density-estimation} are still relevant, CNNs can perform the task well enough without much human intervention. Most recent researches in computer vision have moved to a more advanced task which is semantic segmentation.

\begin{figure}[h]
    \centering
    \includegraphics[width=1\textwidth]{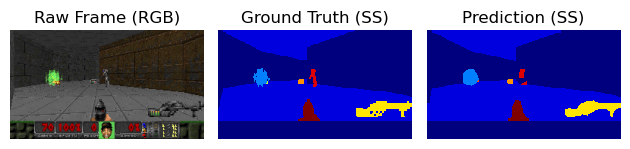}
    \caption{From left to right: an unprocessed RGB game frame from Doom; perfect semantic segmentation result; predicted semantic segmentation by DeepLabV3 model.}
    \label{fig:ss_demo}
\end{figure}

Semantic segmentation (SS) aims to classify \textbf{every pixel} within an input image into one or multiple predefined categories. In a sense, SS can be seen as a further step in image segmentation: instead of segmenting out different objects individually, pixels that belong to objects within the same semantic class are grouped and assigned the same label. Unlike other tasks in computer vision such as image classification, which assigns labels to entire images; or object detection, which identifies and localises certain objects within an image; SS provides information that is essential in understanding the whole scene in a human-like manner. For example in figure \ref{fig:ss_demo}, pixels that belong to walls of different textures are all assigned the same "wall" label since all walls belong to the same semantic class despite the variation in appearances. The pixel-wise classification in SS is crucial to applications that would require a precise interpretation of the environment, which includes many reinforcement learning tasks in fields like robotics \cite{robotics-rl}, autonomous vehicles \cite{autonomous-vehicles-rl} or medical imaging \cite{medical-imaging-rl}.

\subsection{Fully Convolutional Network}
Semantic segmentation has evolved significantly with the rise in popularity of deep neural networks, particularly deep CNNs. Earlier successful approaches made use of fully convolutional networks (FCN) \cite{fcn-ss}, which revolutionised the field by using CNNs for dense prediction tasks. FCN replaced all fully connected layers in typical DNN with convolutional layers, which produces spatial maps of scores for each semantic class which matches the shape of the original input image. This approach yielded good performance when trained in an end-to-end fashion with pixel-wise labelled data acting as the ground truth.

\begin{figure}[h!]
    \centering
    \includegraphics[width=150px]{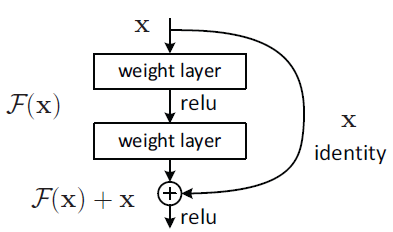}
    \caption{Residual connection, source: figure 2 of \cite{resnet}}
    \label{fig:resblock}
\end{figure}

\subsection{Residual Network}
Residual network (ResNet) \cite{resnet} is a variant of CNN that utilises residual connections (also known as skip connections). As illustrated in figure \ref{fig:resblock}, a residual connection would create a copy of the input to a certain layer inside a CNN, apply a certain transformation to the copy, and add it to the output of a later layer. Two transformations were proposed in the original paper: identity transformation and convolution with a 1$\times$1 kernel, but the former is more commonly used. Residual connections are very effective in avoiding gradient vanishing issues that are common in deep CNNs.

\subsection{Dilated Convolution Kernel}
The receptive field of a convolution kernel is defined as the area in input data that it can use to extract features. For a standard convolution kernel, the receptive field is equivalent to its shape, but a dilated convolution kernel has an additional hyperparameter known as the "dilation rate", which controls the distance between learnable parameters in a dilated kernel.

\begin{figure}[h]
    \centering
    \includegraphics[width=1\textwidth]{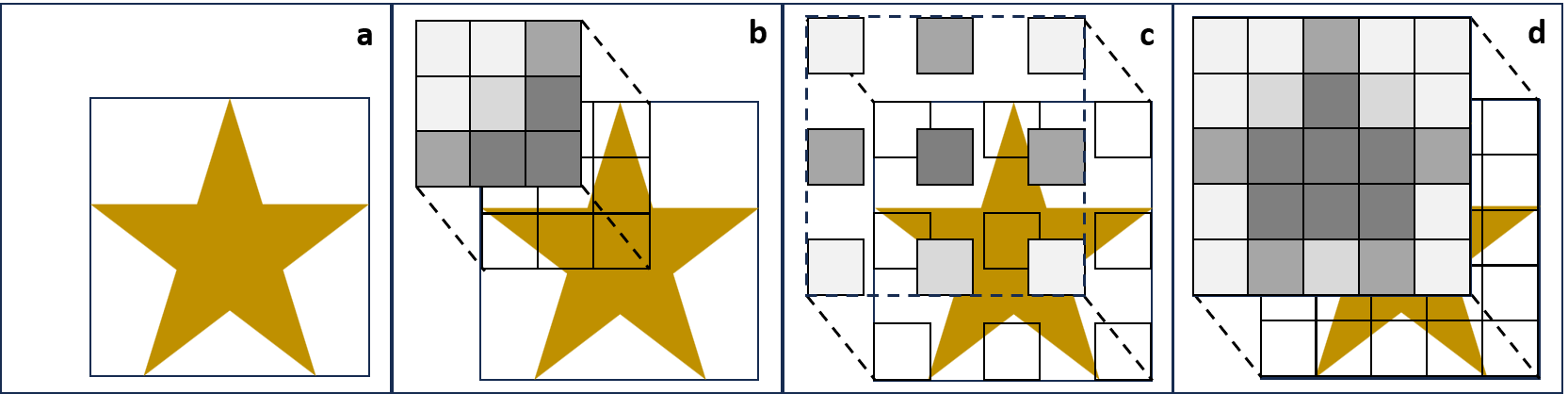}
    \caption{A comparison of feature extraction on a star pattern (a) with: a 3$\times$3 standard kernel (b), a 3$\times$3 dilated kernel (c) with dilation rate = 2, and a 5$\times$5 standard kernel (d).}
    \label{fig:atrous-cmp}
\end{figure}

As demonstrated in figure \ref{fig:atrous-cmp}, a standard 3$\times$3 kernel with 9 learnable parameters is unable to capture the star pattern by itself since it requires a standard 5$\times$5 kernel with 25 learnable parameters. A dilated 3$\times$3 kernel with 9 learnable parameters can act as a compromise to the 5$\times$5 standard kernel and capture the whole pattern with no additional learnable parameters, at the cost of loss in detail.

\subsection{DeepLab}
One of the previous state-of-the-art models in semantic segmentation is DeepLab \cite{DeepLab}, a class of CNN models that demonstrated significant improvements over its predecessors by employing an operation known as \textbf{atrous convolution} or dilated convolution. Atrous convolution is a variant of convolution that expands the receptive field without the introduction of additional learnable parameters by using dilated convolution kernels instead of standard kernels.

\paragraph{Atrous Spatial Pyramid Pooling}
By modifying the stride and dilation rate of dilated kernels, it is possible to capture multi-scale context by adopting different dilation rates in multiple kernels and combining them to form an image pyramid \cite{image-pyramid}. This technique is known as atrous spatial pyramid pooling (ASPP) \cite{DeepLab} and is crucial for the model to capture fine details and understand complex scenes in images. 

\begin{figure}[h]
    \centering
    \includegraphics[width=0.6\textwidth]{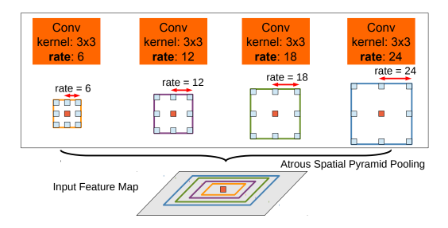}
    \caption{Atrous spatial pyramid pooling, source: figure 4 of \cite{DeepLab}}
    \label{fig:aspp}
\end{figure}

\paragraph{DeepLabV3}
DeepLabV3 \cite{rethinkatrous} is the third iteration of DeepLab and the fastest in inference speed. Compared to its predecessors, DeepLabV3 integrated ASPP with an additional global pooling operation over the feature map to capture image-level features, producing a global feature vector. This vector is then up-sampled and concatenated with the output of atrous convolutions to provide context for the input image as a whole.

\paragraph{DeepLabV3+}
DeepLabV3+ \cite{DeepLabV3+} built upon DeepLabV3 by introducing an encoder-decoder architecture \cite{u-net}. The additional decoder network helps to refine segmentation, especially at object boundaries where DeepLabV3 would sometimes fail to capture fine details. Despite the introduction of techniques such as depthwise separable convolutions \cite{depthwise-separable-convolutions} to reduce the computational cost, it is still slower than DeepLabV3 due to the additional decoder network.

\section{Reinforcement Learning}

Reinforcement learning (RL) is a subfield of machine learning that focuses on finding optimal strategies in various environments. In RL, a decision-maker which is also known as an \textbf{agent} would make observations of the current environment at each time-step, these observations are representations of the current \textbf{states} and the set of all possible states is called the \textbf{state space} of a given problem. The agent would use the policy it has learnt to make decisions based on one or more observations and take actions to which the environment would respond with positive or negative \textbf{rewards}. The agent seeks to maximise cumulative rewards over time by learning an optimal policy, which would dictate the best action to take in each state of the environment. Among the various approaches to solving RL problems, actor-critic methods \cite{actor-critic}, policy gradient methods \cite{policy-gradient}, and proximal policy optimization (PPO) \cite{PPO} which combined these two approaches, have emerged as significant techniques, with PPO models and its recurrent variant used in all of the RL agents trained for this project.

\begin{figure}[h]
    \centering
    \includegraphics[width=0.4\linewidth]{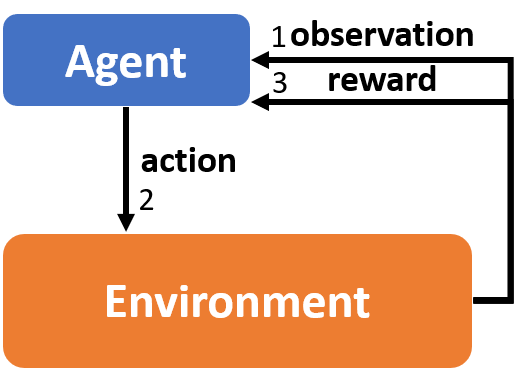}
    \caption{In reinforcement learning, an agent would receive observation of the environment (1), perform a chosen action (2), and receive a reward (3).}
    \label{fig:rl-generl}
\end{figure}

\subsection{Rewards}
Rewards are feedback from the environment given to an agent based on its previous actions, implying whether the actions are desirable given the corresponding states. A positive reward incentivizes the agent to take such actions more often given the same observations, and a negative reward would discourage the agent from learning this state-action pair as it's undesirable. In some environments, positive rewards would be rewards for taking actions that are not directly desirable but may lead to ideal outcomes in the long run, these are called "shaping rewards" \cite{shaping}.

\subsection{Markov Decision Process}
RL problems are often formulated as Markov Decision Processes (MDPs). MDP is a class of optimisation problems where the situation is partially controlled by the decision makers' strategies and partially stochastic. MDPs are discrete-time, meaning the problem is not considered continuous and is modelled with discrete time-steps instead. 

An MDP consists of these following components: a \textbf{state space} $S$ containing all possible states the agent can encounter, a set of all possible actions to take known as the \textbf{action space} $A$, a \textbf{transition function} $T$ that gives the probability distribution of the next possible states given specific state-action pairs, a \textbf{reward function} $R$ determining immediate rewards for transitioning from current state $s$ to another state $s'$. To solve an MDP, the agent must learn an optimal strategy $\pi$ that maximizes a cumulative function for the rewards. The transition functions of MDPs are often unknown in practice, in this case, a simulator model would be used to determine the next state $s'$ given the current state $s$ and chosen action $a$ with a simulation. The current \textbf{policy} $\pi(a_t|s_t)$ of an RL agent determines the action $a_t$ to perform given the current state $s_t$ at time-step $t$.

In MDP, it is assumed that observations of the environment are sufficient to represent the current state. When this assumption does not hold, a partially observable variant of MDP is used to model the problem.

\begin{figure}[h]
    \centering
    \includegraphics[width=0.5\textwidth]{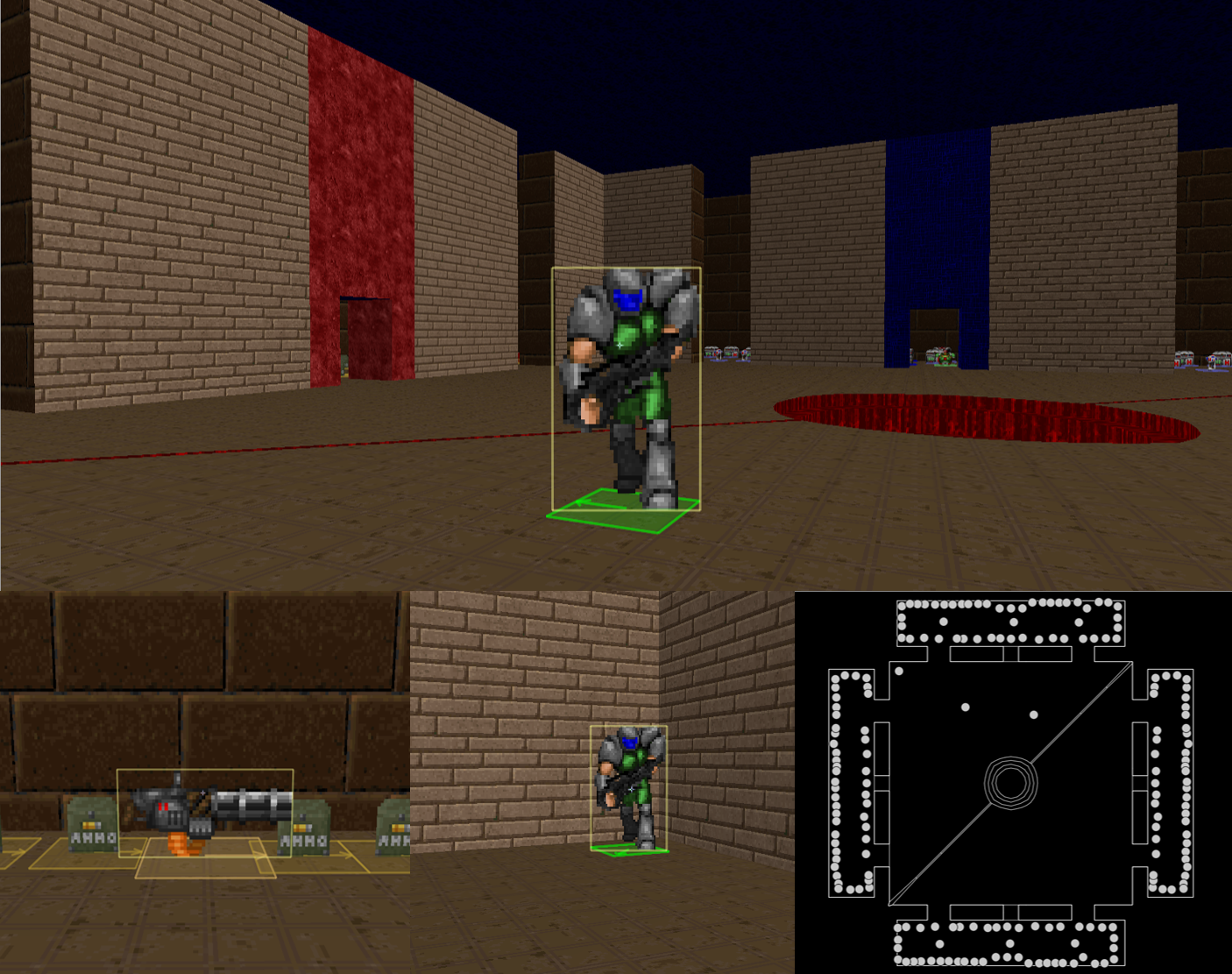}
    \caption{With a 2D view (top), the agent does not have access to some important information, for example: powerful weapons hidden behind walls (bottom-left), enemies outside of viewing angle (bottom-center), or a map of the full environment (bottom-right).}
    \label{fig:2dproj}
\end{figure}

\subsection{Partially Observable Markov Decision Process} 
A partially observable Markov Decision Process (POMDP) \cite{POMDP} is a variation of MDP where the current state cannot be obtained by direct observations: some information crucial to determining the current state is often hidden from the agent in practice. Controlling RL agents in a 3D environment using 2D visual information is often formulated as POMDP \cite{DRQNDoom, game-rl} since the agent only receives 2D projections of a limited view of the whole 3D environment as observations. 

Essential information such as depth is missing and the observations are viewpoint-dependent, with information required to make ideal decisions possibly obfuscated, as illustrated in figure \ref{fig:2dproj}. In POMDP, a state $s_t$ at time-step $t$ is typically represented as the history of observations from $o_{t-k}$ to $o_t$ with a finite length $k$ up until the current time-step $t$. Therefore, models with recurrent architectures such as deep recurrent Q network (DRQN) \cite{DRQN} or recurrent proximal policy optimization (RPPO) \cite{rppo} are often used for agents solving POMDPs for their ability to "memorise" previous observations.

\subsubsection{Frame Stacking}
A common approach to tackle partial observability without the use of a recurrent model is to employ a technique known as frame stacking \cite{DQNAtari}, which simply stacks the current observation with a set number of history observations to form the input to RL agents.



\subsection{Q-Learning}
Q-learning \cite{Q-Learning} is a value-based RL algorithm which learns a state-action value function $Q(s, a)$ that maps a state-action pair $(s, a)$ to its corresponding the Q (quality) value for taking the action $a$ given the current state $s$. This function is known as the Q-function or Q-table. A Q-table has a finite size and needs to store a quality value for every possible state-action pair, meaning that $N_s \times N_a$ entries are required to be reserved in memory when using a Q-table, where the number of possible states and possible actions are denoted by $N_s$ and $N_a$. Traditional Q-learning is not suitable for solving RL problems with a high $N_s$ as a result.


\paragraph{Deep Q Network}
Deep Q network (DQN) is one of the earliest DRL algorithms that were successful in game-playing with raw pixels as input \cite{DQNAtari}, it replaces the Q-table in Q-learning with a neural network. Since the original paper for playing Atari games with DQN using raw pixels as input, DQN has become a popular approach for playing video games using high-dimensional visual input.

\paragraph{Deep Recurrent Q Network}
Deep Recurrent Q Network (DRQN) is a variant of DQN that introduced the recurrent neural network long short-term memory (LSTM) \cite{LSTM} to solve POMDPs more effectively. The previous state-of-the-art \cite{DRQNDoom} in playing Doom with RL utilized a DRQN model for controlling the agent during combat encounters and a DQN model for navigation outside of combat.

\subsection{Actor-Critic}
Instead of learning the optimal policy or value function, actor-critic methods \cite{actor-critic} combined policy-based and value-based approaches to address their respective limitations. In actor-critic methods, an actor model would be responsible for selecting the optimal action based on a policy, while a critic model would estimate the value function and evaluate the chosen actions.

\subsection{Policy Gradient}
Policy gradient methods \cite{policy-gradient} are a class of policy-based algorithms that optimise the agent's policy directly instead of the indirect approach of updating the value function (Q-function in Q-learning). Policy gradient methods are more effective in highly stochastic environments with continuous action spaces compared to value-based approaches such as Q-learning.

In policy gradient models, the current policy $\pi_\theta{(a|s)}$ is parameterised by the current parameters $\theta$ of the model and the objective is to maximise the expected cumulative reward function $J(\theta)$ $w.r.t$ $\theta$. The expected value of $J(\theta)$ is defined as follows, with the initial state denoted by $s_0$:
\begin{equation}
    J(\theta) = V^{\pi_\theta}(s_0)
\end{equation}

The policy gradient theorem defines the gradient of this objective which can be used to perform gradient ascent for optimisation:
\begin{equation}
\nabla_\theta J(\theta) = \mathbb{E} \left[ \nabla_\theta \log \pi_\theta(a|s) Q(s,a) \right]
\end{equation}
In this expression, $Q(s, a)$ represents the action-value function for a given state-action pair. In practice, this Q function is often approximated or replaced by estimators such as the advantage function $A(s, a)$ \cite{advantage-function} or Monte Carlo returns \cite{eps-greedy}, as the true Q values are usually not available.

The term $\log \pi_\theta(a|s)$ guides the weight updates, to increase the log probability of selecting action $a$ given the observation of current state $s$. This approach would stabilise gradient updates and ensure that the optimisation focuses more on actions that are more likely to lead to high returns. By following this gradient ascent, the algorithm searches for a local maximum in $J(\theta)$.

\subsection{Proximal Policy Optimization}
Proximal policy optimization (PPO) \cite{PPO} is a popular actor-critic, policy gradient method that improved on earlier trust region methods like trust region optimization (TRPO) \cite{trpo} by using a less computationally expensive approach for maintaining stability during training. PPO optimises a surrogate objective that constrains the update step to prevent large deviations between a new policy and the current one, maintaining its simplicity while incorporating the trust region techniques from TRPO with the use of a clipping function. The objective function of PPO is defined as follows:
\begin{equation}
    L^{PPO}(\theta) = \mathbb{E}_t \left[ \min\left( r_t(\theta) A_t, \text{clip}\left( r_t(\theta), 1 - \epsilon, 1 + \epsilon \right) A_t \right) \right] 
\end{equation} where $r_t(\theta) = \frac{\pi_\theta(a_t | s_t)}{\pi_{\theta_{\text{old}}}(a_t | s_t)}$ is the probability ratio that compares the probability of taking action $a_t$ given observation of the current state $s_t$, under the current policy parameterised by $\theta$ and the old policy $\theta_{\text{old}}$; $A_t$ is the value of the advantage function \cite{advantage-function} at current time-step $t$, a measurement of how much better $a_t$ is compared to the average expected value of actions in state $s_t$ under $\theta$; $\epsilon$ is a hyperparameter that controls to what extend $\theta$ is allowed to deviate from $\theta_{\text{old}}$, typically set to a small value like 0.2. 

The two terms within the $\min$ function reflect the unclipped objective $r_t(\theta) A_t$ and clipped objective $\text{clip}\left( r_t(\theta), 1 - \epsilon, 1 + \epsilon \right) A_t$, the clipping mechanism strikes a good balance between exploration and training stability, it is the main reason PPO is known to be stable \cite{DRL-that-matters} without the need for extensive hyperparameter tuning.

\paragraph{Recurrent Proximal Policy Optimization}
Similar to the relationship between DRQN and DQN, recurrent proximal policy optimization (RPPO) \cite{rppo} is a variant of PPO that introduced LSTM to add recurrency and solve POMDPs more effectively. However, it is less stable from our experiences and PPO that utilises frame-stacking is usually able to yield comparable performance to RPPO \cite{ppo-vs-rppo}.

\section{Compression Potential of Semantic Segmentation as Input Representation in Reinforcement Learning}
Semantic segmentation results are naturally more compressible than RGB colour images. RGB images in 3D environments are often very noisy since they capture a discretised version of the full spectrum of colour information in the scene, including frequent variations in colour due to slightly different lighting conditions. SS masks in contrast only record the semantic class of each pixel in the scene, a discrete value with small and limited range. 

In high-quality SS results, pixels of the same semantic class are typically grouped spatially close to each other, forming large and contiguous regions (as demonstrated in figure \ref{fig:ss_demo} where the majority of the screen is covered in 3 contiguous areas of ceiling, walls and floor). This characteristic allows lossless\footnote{Lossless compression allows the original information to be uncompressed without any modification.} compression techniques like Huffman coding \cite{huffman}, Lempel-Ziv-Welch (LZW) encoding \cite{LZW} and run-length encoding (RLE) \cite{run-length-encoding} to be applied effectively. 

\subsection{Suitability of Common Lossless Compression Techniques}
Huffman coding is a method that requires near-perfect knowledge of the distribution of symbols (pixel values in the case of images) beforehand, which is often not practical in RL, where the distribution changes frequently with the current policy of the agent. For example, an agent may initially learn a policy that hugs walls most of the time, but as it moves to new policies that try to accomplish the task instead of getting stuck at walls, the frequency of seeing walls would decrease significantly. A frequent update for the definition of symbols (also known as the Huffman tree) in Huffman coding would induce big overheads, significantly slowing down the training of RL agents.

LZW would work well for the task, but it is not easily vectorisable in its common form, as the operation of LZW is highly dependent on the previous data processed and this sequential workflow along with its variable output size limited its ability to be parallelised with modern multi-core hardware.

RLE is a suitable algorithm for the task, despite its simplicity. RLE simply identifies repeating symbols or patterns in the raw data and replaces them with runs\footnote{Repeated sequence of the same symbol or pattern.} of the symbol or pattern. The identification of runs with a specific symbol is not dependent on the identification with another symbol, which makes RLE very vectorisable. 

\section{AI for Video Games}
Learning to play video games is a relatively popular task in the field of visual-perception-based RL, many algorithms that were developed for this task such as DQN \cite{DQNAtari} and DRQN \cite{DRQN} have proven to generalize well to other areas in reinforcement learning like robotics \cite{DRQNRobotics}. Video games such as the Atari 2600 games \cite{atari-env}, Doom \cite{ViZDoom} and Unreal Tournaments \cite{pogamut} have been naturally suitable as benchmarks for artificial intelligence (AI) algorithms due to the ease of manipulation and well-designed rule sets that have been tested by gamers around the world due to their commercial origins. First-person shooter games in particular are suitable for testing algorithms that can be adapted to more practical fields such as robotics.

The earliest published work in training AI for first-person shooter games \cite{rl-game-2002} focused on modelling human player behaviours in the game Soldier of Fortune 2, later works in this direction turned to the Unreal Tournament (UT) series due to the convenience provided by POGAMUT \cite{pogamut}, a middle-ware platform that communicates with UT to allow for controlling in-game bots with AI algorithms. These works while fascinating, do not transfer well to real-world scenarios due to the input data being relatively high-level and abstract, often containing information that is hidden from the AI agents and would need to be inferred in real-world scenarios. Most recent works utilise high-dimensional sensory inputs and are consequently closer to practical use \cite{game-rl}.

Other interesting applications for AI in video games include: emulation of the Doom game engine \cite{game-n-gen} via diffusion models \cite{diffusion}, creation of digital humans via generative AI \cite{nvidia-ace} to act as NPCs\footnote{Non-playable characters in a video game.} in-game, and automated character creation with the Unity game engine powered by large language models \cite{llm-unity}.

\subsection{Doom}
Originally released in 1993, Doom is a 3D First-Person-Shooting (FPS) video game that revolutionised the video game industry. Despite its controversies in video game violence and addictive game-play in the 1990s, Doom has become one of the most memorable works in the history of video games and inspired many works in the field of AI, including RL agents that learnt to play its deathmatches \cite{ViZDoom, DRQNDoom, FYP, PreviousSS}, diffusion models that emulate its graphics and game logic \cite{game-n-gen} and semantic segmentation models that learn to segment its game frames \cite{doom-dataset-hakan}.

\subsection{ViZDoom}
ViZDoom \cite{ViZDoom} is an open-source project designed specifically for training RL agents that play Doom with purely visual information as input. ViZDoom was based on ZDoom \cite{zdoom}, a source port of Doom's game engine and its API provided direct access to the game engine, enabling many useful features including: 
\begin{enumerate}
    \item The use of custom maps with customisable textures, enemy behaviour, etc.
    \item Multiple options for rendering the game, including colour modes, resolutions, whether to render certain in-game elements, etc.
    \item Activation of console commands and cheat codes during game-play.
    \item Access to internal variables that can be defined in ACS scripts for custom maps.
    \item Access to information of game objects that are rendered on-screen and a labels buffer which labels every pixel on-screen with an id of the object it belongs to.
\end{enumerate}

Feature 5 in particular allowed perfect semantic segmentation results of a raw RGB game frame to be extracted as the ground truth for training SS models.

\subsubsection{Creating Custom Maps}
To create a custom map (also known as "scenarios" in ViZDoom) for training RL agents, a map editor software like SLADE3 \cite{SLADE3} can be used, which would provide a graphical user interface for editing map layouts and textures. Customised scripted events can be added with a C-like scripting language known as action code script (ACS). ACS scripting allows custom reward definitions and events such as giving shotguns to all players or opening a specific door after every enemy has died.

\subsection{Frame-skipping}
Frame-skipping \cite{atari-env} is a technique widely adopted in previous approaches \cite{atari-env, DRQNDoom, FYP, other-FYP-ppo, PreviousSS} for training RL agents to play video games, where the RL agents only receive input observation every $k$ time-steps (frames), with the chosen action from the agents repeated over all of the skipped frames. Frame-skipping is a common practice in ViZDoom-related literature \cite{DRQNDoom, FYP, other-FYP-ppo, PreviousSS}, with $k=4$ \cite{DRQNDoom} widely accepted to be the best value for ViZDoom in general.

\chapter{Methodology \& Evaluation Framework}

\section{Conceptual Design}
This section outlines the conceptual design of the project, focusing on the integration of semantic segmentation into reinforcement learning (RL) agents operating in 3D environments. Our primary goal was to investigate how different input representations: raw RGB images, semantic segmentation (SS) masks, and a combination of both would affect the performance and memory efficiency of RL agents. Three representations of a Doom game frame have been analysed in this project, with the well-adopted RGB input \cite{ViZDoom, DRQNDoom, PreviousSS, FYP, other-FYP-ppo} acting as our baseline:

\begin{enumerate}
    \item \textbf{RGB (baseline)}: the most common input representation for RL in 3D environments, known to yield better performance over grayscale images \cite{DRQNDoom}.
    \item \textbf{SS-only}: a novel representation that utilises DNN-predicted SS masks as input to RL agents, capable of reducing memory consumption of RGB by 66.6\% without a significant impact on agents' performance. With a vectorised run-length encoding (RLE) compression, the memory consumption can be further optimised to less than 2\% of RGB without much overhead.
    \item \textbf{SS+RGB}: a novel representation that adds a DNN-predicted SS mask as an additional colour channel to augment the RGB image and provide additional semantic information to improve the performance of RL agents.
\end{enumerate}

We decided to test these three input representations in custom maps via the ViZDoom platform and created a framework for training deep reinforcement learning agents with semantic segmentation to play Doom deathmatches against built-in bots.

\begin{figure}[h]
    \centering
    \includegraphics[width=0.5\linewidth]{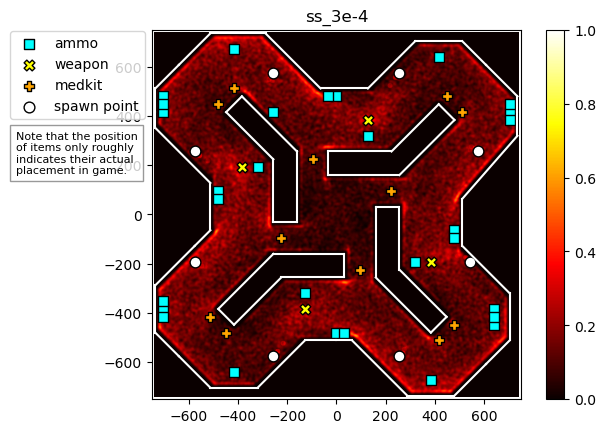}
    \caption{A positional heatmap of the RL agent used to gather data for training SS models, an SS-only PPO agent trained with perfect SS input.}
    \label{fig:collect-ss-agent}
\end{figure}

Our workflow consists of the following steps:
\begin{enumerate}
    \item Select a map for training both the SS model and the RL agents.
    \item Train an initial batch of RL agents on the selected map using different hyperparameters (only learning rate for this project due to time constraints) and different representations of input visual data: RGB (baseline), SS-only, SS+RGB. 
    \item Pick the best-performing agent of each input representation for further evaluation (performance should be measured by the in-game score: frags, instead of rewards).
    \item Collect positional data from evaluation episodes and perform heatmap analysis for agents' movement to pick the most suitable agent (that visits every corner of the map most uniformly) for collecting labelled semantic segmentation data. The heatmap for our data collection agent is shown in figure \ref{fig:collect-ss-agent}.
    \item Run the data collection agent in the selected map and collect labelled game frames (RGB game frames + perfect semantic segmentation result from ViZDoom) from a large number of evaluation episodes (200 in our case).
    \item Train at least one DNN model for semantic segmentation.
    \item Perform further hyperparameter search (as SS-only and SS+RGB have different learning rate requirements than RGB according to our testing).
    \item Evaluate the trained RL agents on both seen and unseen maps with perfect semantic segmentation information (for applicable agents) and with DNN-predicted semantic segmentation results. The unseen maps should by default use similar textures for walls, floors and ceilings, but alternative versions that use different textures can be evaluated as well. 
    \item Finally, evaluate the performance of built-in bots that utilised node-based navigation on the maps by hosting bot-only episodes and recording the highest scores in each episode, this can be used as a non-neural baseline.
\end{enumerate}

\section{Selected Maps}

\begin{figure}[h]
  \centering
  \begin{subfigure}[b]{0.5\linewidth}
    \includegraphics[width=\linewidth]{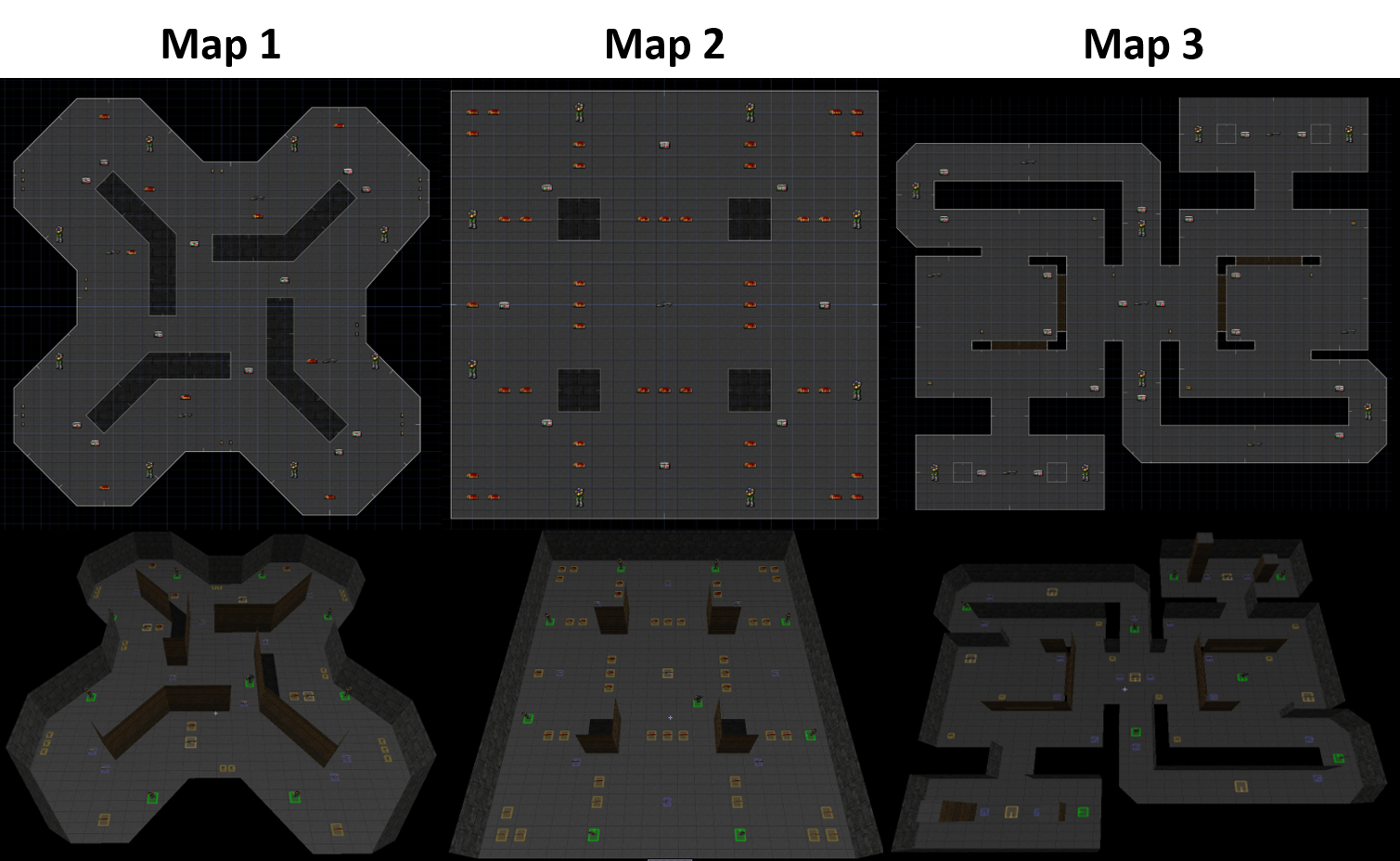}
    \caption{2D (top) and 3D (bottom) layouts of the three maps, as displayed in SLADE3 map editor \cite{SLADE3}.}
  \end{subfigure}
  \begin{subfigure}[b]{0.4\linewidth}
    \includegraphics[width=\linewidth]{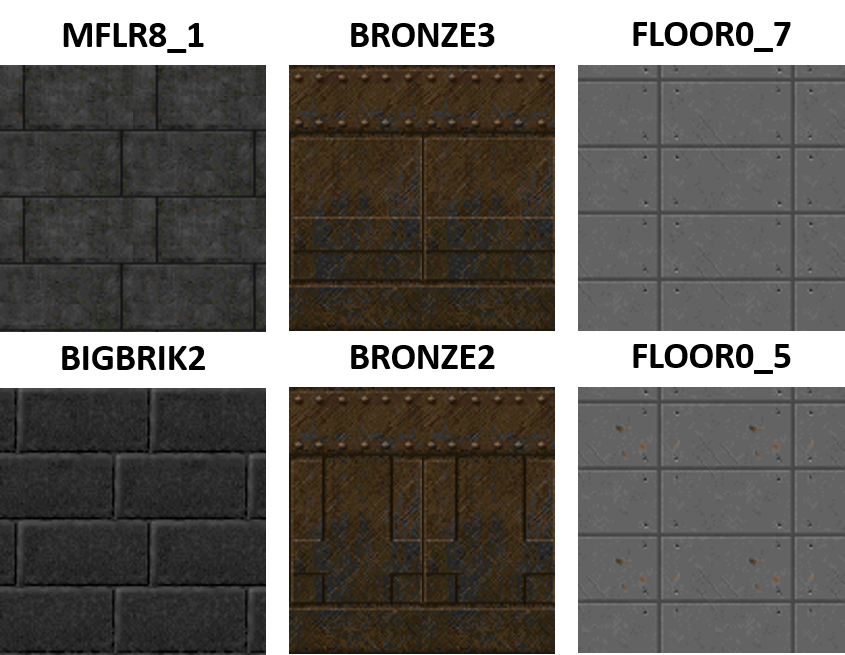}
    \caption{Original textures in Map 1 (top) and alternatives for 
     comparison (bottom).}
  \end{subfigure}
  \caption{Selected maps and alternative textures.}
  \label{fig:maps-fig}
\end{figure}

We gathered three custom maps from an open-source project \cite{doom-maps}\footnote{\href{https://github.com/lkiel/rl-doom}{https://github.com/lkiel/rl-doom}}\footnote{Slight modifications are made on Map 2, a pistol-only scenario, to give players access to shotguns.} for deathmatches in Doom, a game-play mode that puts multiple players and possibly bots on the same map, counting the number of kills within a predefined time limit as scores (known as "frags" in Doom). Each map selected, as demonstrated in figure \ref{fig:maps-fig} and table \ref{tab:map-characteristics} is representative of a certain type of game-play situation to our best effort. \textbf{The training of RL agents and the SS model are all performed on Map 1 only.} This is for comparability consideration of our results since Map 1 was used to train/evaluate RL agents and SS model in the previous study \cite{PreviousSS} for applying semantic segmentation to ViZDoom deathmatches against built-in bots. 

\begin{table}[h]
\centering
\begin{tabular}{|l|c|c|}
\hline
Map & Navigation Complexity & Combat Intensity\\
\hline
Map 1 & Medium & Medium \\
\hline
Map 2   & Low & High \\
\hline
Map 3 & High & Low \\
\hline
\end{tabular}
\caption{Characteristics of the three maps.}
\label{tab:map-characteristics}
\end{table}

\label{changing-texture}
In addition to these three maps, two additional variants of Map 1 have been tested, as illustrated in figure \ref{fig:map1-cmp} of the appendix. The first one would alter the wall textures (MFLR8\_1 and BRONZE3) and another would additionally add moss to the floor texture (FLOOR0\_7), as shown in part (b) of figure \ref{fig:maps-fig}\footnote{Textures tiled for better visualisation, all of the textures come from the Freedoom project \cite{freedoom}.}.

\section{ViZDoom Configurations}
For all of our maps, we used the following configurations:
\begin{itemize}
    \item episode\_timeout = 5250 (2 minutes and 30 seconds, default to these maps)
    \item death\_penalty = 0 (this penalty needs to be disabled in multi-player)
    \item living\_reward = 0 (for deathmatches, we don't incentivise survival directly)
    \item render\_hud = true (a good idea to disable, but kept for human readability)
    \item render\_crosshair = false (might introduce complexity for semantic segmentation)
    \item render\_weapon = true (allow agents to be aware of the weapon they're holding)
    \item render\_decals = false (no reason to keep, disable to decrease rendering cost)
    \item render\_particles = false (no reason to keep, disable to decrease rendering cost)
\end{itemize}
A screen resolution of 256$\times$144 was used for all of our agents, as it was the lowest resolution with a 16:9 aspect ratio, providing a $108^{\circ}$ field of view (FOV) and more available information compared to 4:3 with only $90^{\circ}$ FOV \cite{DRQNDoom}. We enabled labels buffer access to obtain perfect semantic segmentation results and off-screen rendering to avoid the additional overhead between the software renderer of ViZDoom and the operating system. As mentioned in section \ref{avoid-down-sampling} in our introduction, we would not be utilising down-sampling due to the hidden assumptions made when choosing down-sampling techniques for SS results. However, we do believe that a comparison of down-sampling algorithms would be interesting for future studies.

\subsection{Action Space}
Every RL agent is allowed access to the following buttons: 

(1) ATTACK, (2) MOVE\_FORWARD, (3) MOVE\_LEFT, (4) MOVE\_RIGHT, (5) TURN\_RIGHT, (6) TURN\_LEFT.

This creates a relatively large action space with $2^6=64$ possible combinations of buttons. To limit the number of actions, we took the approach from literature \cite{PreviousSS} and filtered out nonsensical combinations such as turning/moving left and right simultaneously. Additionally, all combinations that perform any other action with an attack were removed to simplify the decision process. After also removing the idling action (not pressing any button), the size of our action space was reduced to 18.

\subsection{Additional Game Arguments \& Bots}
Each game-play episode is initialised with several additional game arguments\footnote{Information on game arguments can be found here: \href{https://zdoom.org/wiki/CVARs:Configuration}{https://zdoom.org/wiki/CVARs:Configuration}}:

\verb|-host 1 -deathmatch +viz_nocheat 0 +cl_run 1 +name AGENT| would host a deathmatch, enable access to buffers considered as cheating in ViZDoom, including the labels buffer needed for obtaining perfect SS results, force all players to always run and set up the name of the current player.

\verb|+sv_forcerespawn 1 +sv_respawnprotect 1 +sv_nocrouch 1 +sv_noexit 1| would force dead players to respawn automatically, ensure that players are immortal for a few frames after respawning, disable crouching, and disable exiting for the map.

For bot-only matches, the \verb|+sv_cheats 1| argument is also required, as the player needs to be protected with the \verb|iddqd| cheat code to ensure that it does not get killed by gunshots from bots not aiming at it. The command \verb|bot_observer 1| would also be sent to the game engine for bots to ignore the observing player.

A total of 8 bots are added at the start of each game-play episode of RL agents via the \verb|addbot| command, a ninth bot would be added to bot-only matches to keep a consistent player count of 9. The default bot configurations from ZDoom \cite{zdoom} are used. 

\section{Labels Buffer in ViZDoom \& Semantic Segmentation}
\label{vizdoom-ss-labels}
When the labels buffer is enabled in ViZDoom, a list of rendered objects and a labels buffer will be accessible. Inside the labels buffer are the unique IDs of the front-most object occupying each pixel and the list of rendered objects also contains the name and unique IDs for each object. By combining these two information, we can correctly label the semantic class of most pixels.

\subsection{Exceptions}
There are three exceptions to the described method: walls, UI elements and the controllable player itself. The walls are considered line definitions instead of objects within the rendering logic of Doom and automatically get a unique ID of 1 in the labels buffer; UI elements can simply be ignored and not segmented since they conveniently form a rectangular shape occupying the entirety of row 121 to 144 in the frame buffer of our current resolution; the controllable player, after testing, was discovered to always appear at the end of the rendered objects (as it needs to be rendered just below UI elements) allowing us to identify its presence at a low computational cost.

\subsection{Semantic Classes}
Objects in the scene are grouped into the following semantic classes, same as \cite{PreviousSS}: (1) Floor/Ceiling, (2) Wall, (3) ItemFog, (4) TeleportFog, (5) BulletPuff, (6) Blood, (7) Clip, (8) ShellBox, (9) Shotgun, (10) Medikit, (11) DeadDoomPlayer, (12) DoomPlayer, (13) Self. Objects with unknown semantic classes are labelled as Floor/Ceiling.
We did not experiment with alternative definitions of semantic classes due to time constraints.

\section{Run-Length Encoding Applied to SS-Only Inputs}
Traditionally, run-length encoding (RLE) stores the repeated symbol and length of each run in the same data stream/array, this requires the symbol and maximum length of a run to share the same bit-depth. In our case where the number of unique symbols is smaller than the length of the uncompressed data, the additional bits used to store each symbol hold no meaningful information and are essentially wasted. An improvement, in this specific case, would be to save RLE-compressed data in three arrays of different bit-depth, storing the repeated symbol/starting position/length of every run separately. 

This variant, as described in algorithm \ref{alg:vec_rle} of the appendix, has a vectorised implementation \cite{numpy-rle} that was incorporated into this project for evaluation of memory consumption after switching the input to RL agents from RGB to the proposed SS-only representation. 

\section{Reinforcement Learning Environment}
\subsection{Environment Wrapper}
To allow customisation in various aspects of the map, we created an OpenAI Gymnasium  \cite{gymnasium} wrapper for ViZDoom with support for features such as customisable reward calculation, frame stacking and recording of game-play episodes in RGB+SS representation. For all of our experiments, we utilised the frame-skipping technique from literature \cite{atari-env, DRQNDoom, PreviousSS, FYP, other-FYP-ppo} and set the frame-skip to 4, an optimal value that strikes a good balance between training speed and action precision \cite{DRQNDoom}. With this setting, the RL agents would receive observations every 4 time-steps and decide on the action to repeat before the next observation.

\subsubsection{Rewards}
The definition of rewards as shown in table \ref{tab:reward-def}, is kept the same as in reference works \cite{doom-maps, PreviousSS} that used the same maps. For the detection of movement, a topic not covered in the reference works, we measured the Euclidean distance between positions of subsequent updates and penalised the agent if displacement was less than 3 units. This value is specifically picked for our setup with a frame-skip of 4, as Doom players have a maximum frictionless acceleration of 1.5625 units per $\text{tic}^2$ \cite{doom-movement}\footnote{Tic is a single time-step in Doom's game logic.} and speed limit of 30 per tic. We believe that setting the displacement threshold to 3 units strikes a good balance between incentivising movement and not encouraging reckless strategies.

\begin{table}[h]
    \centering
    \begin{tabular}{|l|c|l|c|}
    \hline
        Get frag & +1 & Death & -1\\
    \hline
        Gain x health points & +0.02x & Lose x health points & -0.01x\\
    \hline
        Gain x units of ammo & +0.02x & Spend x units of ammo & -0.01x\\
    \hline
        Moved after last update & +0.00005 & Not moved after last update & -0.0025\\
    \hline
        Damage enemy by x health points & +0.01x & & \\
    \hline
    \end{tabular}
    \caption{Reward definition.}
    \label{tab:reward-def}
\end{table}

\begin{figure}[h!]
    \centering
    \includegraphics[width=0.6\textwidth]{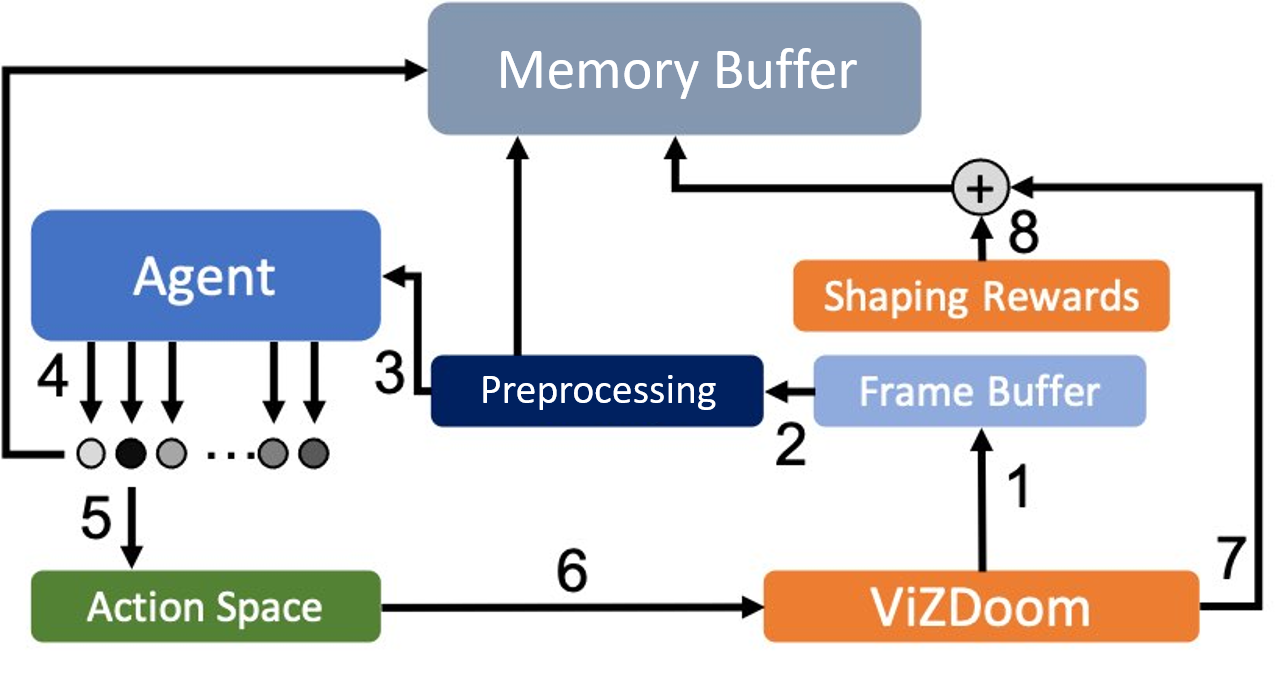}
    \caption{A data-flow diagram of RL in ViZDoom, modified from figure 2 of \cite{FYP}}
    \label{fig:rl-in-doom}
\end{figure}

\subsection{Training Procedures}
As illustrated in figure \ref{fig:rl-in-doom}, the training of RL agents within a ViZDoom environment has the following steps:
\begin{enumerate}
    \item ViZDoom renders the current frame into its frame buffer.
    \item Content of the frame buffer is pre-processed with custom algorithms for resizing or feature extraction (semantic segmentation in our case).
    \item An agent receives the pre-processed input observation and decides on an action to perform.
    \item The selected action and current observation are stored in a memory buffer.
    \item The selected action is translated into valid button and mouse combinations.
    \item The action is sent to ViZDoom and game logic advances for one step.
    \item ViZDoom returns a reward calculated from internal game engine variables.
    \item Additional reward shaping would optionally be applied to add shaping rewards.
    \item The total reward is also stored in the memory buffer.
\end{enumerate}
The memory buffer would be called replay buffer in RL agents that utilise experience replay and rollout buffer in proximal policy optimization (PPO), serving slightly different purposes. In our case with PPO-based RL agents, the rollout buffer would collect multiple trajectories (sequences of state-action pairs) before making updates to the policy networks in PPO. This aggregation ensures that every update update is based on more diverse experiences, reducing the chance of overfitting to specific trajectories.

\section{PPO-based Reinforcement Learning Agents}
The PPO agents in this project are implemented with tools provided by the main and development branches of the open-source Stable Baselines3 (SB3) project \cite{SB3}, utilising PPO models with CNN-based feature extractors. SB3 uses the OpenAI Gymnasium \cite{gymnasium} standard for training \& evaluation environments, providing a unified API structure for various state-of-the-art RL algorithms, allowing the PPO-based agents to train on vectorised versions of our custom environments utilising multi-processing capabilities of our training hardware\footnote{64GB DDR4 RAM, i9-12900HX CPU, RTX 3080Ti Mobile GPU with 16GB GDDR6 VRAM}.

\begin{figure}[h]
    \centering
    \includegraphics[width=1.0\textwidth]{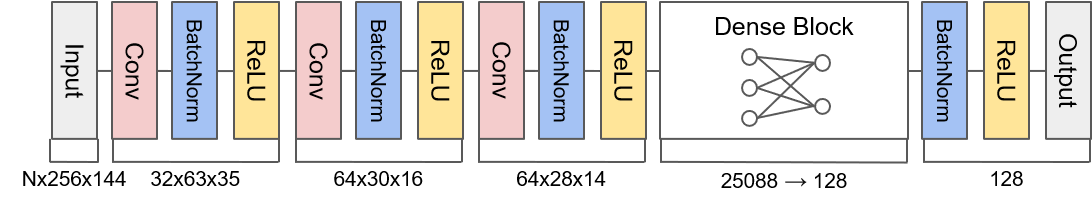}
    \caption{The feature extractor CNN for our RL agents. \textbf{N} refers to the number of colour channels in the input representation.}
    \label{fig:feat-extract}
\end{figure}

As illustrated in figure \ref{fig:feat-extract}, our feature extractor CNN uses a similar architecture as the previous literature \cite{PreviousSS}, with three convolution blocks. Each block consists of a 2D convolutional layer, a batch normalisation layer \cite{batch-norm} and a ReLU activation function applied to the normalised output. The convolution blocks have kernel sizes of 5/4/3 and strides of 4/2/1 respectively, extracting high-level features from the input data. A fully connected layer is inserted after the convolution components to flatten the output feature maps and condense them into a 128-dimensional feature vector, batch normalisation and ReLU are applied to the feature vector to produce the final output.

A simple 2-layer ANN with tanh (hyperbolic tangent) activation function and 64 neurons in each hidden layer would take the extracted 128-dimensional feature vector as input and predict the optimal action to take with current observations.

\paragraph{Justification}
The choice of ReLU activation function in the CNN is supported by evidence \cite{activation-functions} that showed the unsuitability of other common activation functions like logistic sigmoid and tanh in CNN models. The 2-layer ANN architecture was considered a standard baseline model in SB3 and our results showed that it had enough capacity for playing the selected maps using visual inputs, with our best RGB+SS agent yielding performance nearly $2\times$ of the best built-in bots\footnote{The ZCajun bots were considered human-like for its latest version, with a player quoting "Instead of shooting fish in a barrel, you become the fish.": \href{https://www.doomworld.com/mellow/bots.shtml}{https://www.doomworld.com/mellow/bots.shtml}} and $3\times$ of the author\footnote{With a resolution of 1920$\times$1080 and refresh rate of 35 frames per second, the \textbf{best} score we obtained with the same buttons available in Map 1 was 8 (mean was about 5), while the RGB+SS agent yielded a \textbf{mean} score of 19.2 with real-time semantic segmentation.}.

\subsection{Input Representations}
All of our RL agents were trained with perfect semantic segmentation results extracted with the method described in section \ref{vizdoom-ss-labels}, if applicable. Different RL agents would utilise different representations of the input observations:
\begin{enumerate}
    \item \textbf{RGB (baseline)}: raw content of the frame buffer, with 3 colour channels.
    \item \textbf{SS-only}: a semantic segmentation mask, with 1 colour channel.
    \item \textbf{RGB+SS}: the 3 channels of \textbf{RGB} is stacked with the additional channel from \textbf{SS-only} to form a 4-channel input.
    \item \textbf{SS(4)}: a frame-stacking variant of \textbf{SS-only}, where the 1-channel \textbf{SS-only} input of the current observation is stacked with \textbf{SS-only} inputs from the last 3 time-steps to form a 4-channel input.
\end{enumerate}
\textbf{SS(4)} has been added as an alternative to \textbf{RGB+SS}, with the same 4 channels of input. We hypothesised that the short history of observations would allow the SS(4) agents to perform better at navigation tasks in complex scenarios such as the corridors of Map 3. To accommodate for the increase in complexity involved in taking a history of observations as input, we doubled the width of the ANN in SS(4) agents to 128.

\subsection{Hyperparameters}
The hyperparameters for the PPO-based RL agents utilising RGB, SS-only and RGB+SS as input representations, are kept consistent except for learning rates. The agents would interact with 4 vectorized environments (venvs), generating fixed-length trajectories of 4096 time-steps. During each training iteration, $N = 4$ parallel actor models would be collecting trajectories, resulting in data from $N \times T = 4 \times 4096 = 16384$ time-steps to be stored within the rollout buffer. The surrogate loss is subsequently computed using the data in the rollout buffer, and the agents are optimised over $K = 3$ epochs with a batch size of 32. For SS(4) agents, the batch size is doubled to stabilise training. 

\emph{All RL agents are trained for a total of 4 million time-steps, 1 million steps per venv.}

\subsection{Failed Side Experiments with Recurrent PPO}
We also experimented with recurrent proximal policy optimisation (RPPO). We planned to conduct multiple experiments with RPPOs, but all of our setups with various network architectures and hyperparameters failed with catastrophic forgetting. The only one that worked to some degree within the time constraints of this project was an RPPO-based RL agent that replaced the ANN in PPO with a one-layer long short-term memory (LSTM) model with a width of 64. This agent would share the LSTM model between actor and critic, which we suspected is the main reason behind its training stability.

\section{Heatmap Analysis for RL Agents' Movement}
With our custom wrapper for ViZDoom, positional data at each time-step was collected for every evaluation run, allowing us to plot heatmaps to visualise each RL agent's movement pattern. We adopted a density-based heatmapping approach and for the positional data at each time-step, we colour a circle with a radius of 10. The intensity at the centre is set to 10 with a linear or exponential decay in intensity as distance from the centre increases. After experimenting with different colouring schemes, we decided to use this simple equation for its low complexity and ability to highlight small details:
\begin{equation}
    intensity(distance) = max(0, radius - distance)
\end{equation}

\section{Creation of Custom Semantic Segmentation Dataset}
There are existing semantic segmentation (SS) datasets for Doom like the CocoDoom dataset \cite{doom-dataset-hakan}, however, these works were mostly done with copyrighted content (maps and assets) of the original Doom game. Like most ViZDoom-based works \cite{ViZDoom, DRQNDoom, FYP, other-FYP-ppo, PreviousSS, doom-maps}, our project made use of royalty-free assets from the Freedoom project \cite{freedoom}, requiring custom datasets to be built for semantic segmentation.

As mentioned in section \ref{vizdoom-ss-labels}, we utilised the labels buffer and list of rendered objects provided by ViZDoom to extract perfect semantic segmentation results from game-play sessions of RL agents. This allowed us to build a dataset for semantic segmentation in Map 1 by recording 200 game-play episodes of a specific RL agent: an SS-only agent trained at a learning rate of 3e-4 with perfect SS results, the heatmap of which has been displayed in figure \ref{fig:collect-ss-agent}. A total of 247,068 frames were collected at a size of 256$\times$144.

\section{Real-time Semantic Segmentation with DeepLabV3}
For training a model to perform real-time semantic segmentation (SS), we split the whole dataset randomly into training and validation sets with a 9:1 ratio. As mentioned in previous research \cite{SSMario}, the DeepLabV3 model from PyTorch \cite{PyTorch} can be initialised with weights pre-trained on a subset of the COCO 2017 dataset \cite{ms-coco}, which yielded better performance than models trained from scratch. Although the previous work focused on training an SS model for segmenting game frames of Super Mario Bros, a 2D sprite-based platforming game, we decided that it made sense to assume that SS on Doom's sprite-based 3D graphics could also benefit from the pre-training.

\begin{figure}[h]
    \centering
    \includegraphics[width=1\linewidth]{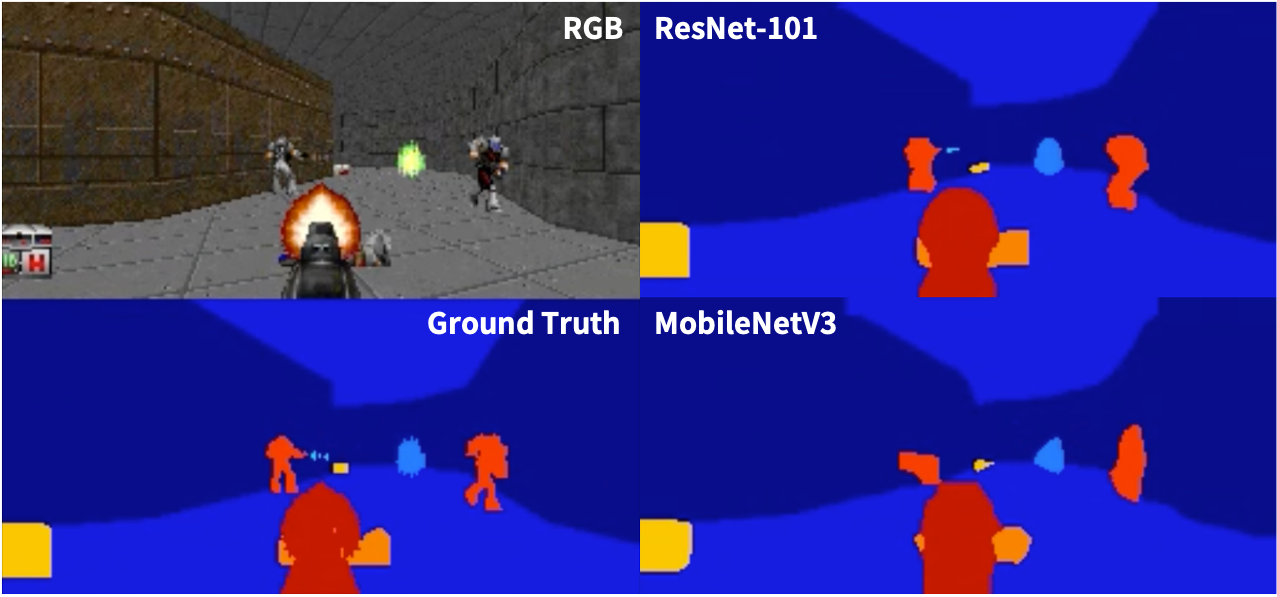}
    \caption{A comparison of semantic segmentation results with different backbones.}
    \label{fig:cmp-deeplabv3-backbones}
\end{figure}

To train the DeepLabV3 model with a different number of semantic classes from the 20 of the COCO dataset, we replaced the final classifier layer with a near-identical 2D convolutional layer that had 13 output channels instead of 20. As demonstrated in figure \ref{fig:cmp-deeplabv3-backbones}, both ResNet-101 and MobileNetV3 \cite{mobilenetv3} were tested as the backbone to DeepLabV3, but the MobileNetV3 model appeared to have insufficient capacity for this specific task and had difficulty yielding an mIoU score greater than 0.65 during training.\footnote{Full comparison video of ResNet-101 and MobileNetV3 as backbones for real-time semantic segmentation can be viewed on the GitHub repository of this project: \href{https://github.com/Trenza1ore/SegDoom}{https://github.com/Trenza1ore/SegDoom}} It was clear that MobileNetV3 learnt to localise objects in the scene with a certain degree of success, but it failed to capture the fine shapes of objects, resulting in random-looking polygons. ResNet-101 was chosen as the encoder backbone of our final DeepLabV3 model for its ability to capture an object's outline properly.

\chapter{Results \& Evaluation}
\textbf{Note:} the heatmaps could not fit due to page limits and are kept in appendix instead.
\section{Performance of RL Agents}
\begin{figure}[h!]
  \centering
  \begin{subfigure}[b]{1.0\linewidth}
    \includegraphics[width=\linewidth]{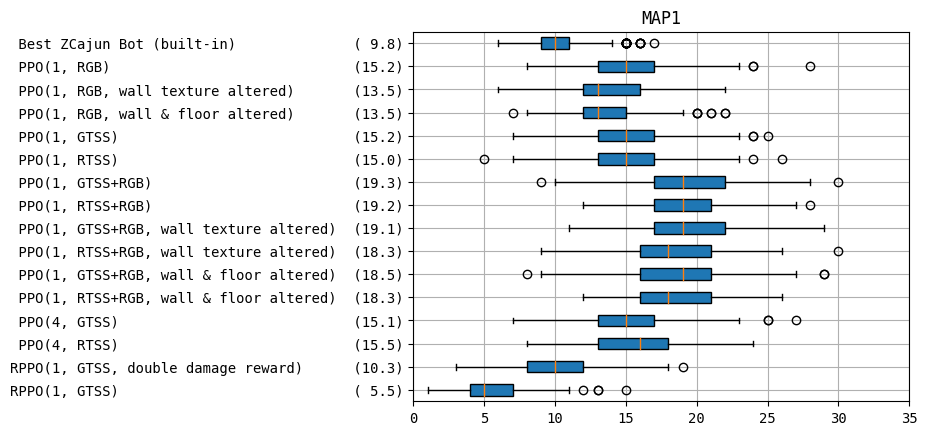}
    \caption{Evaluation Performance on Map 1.}
  \end{subfigure}
  \begin{subfigure}[b]{1.0\linewidth}
    \includegraphics[width=\linewidth]{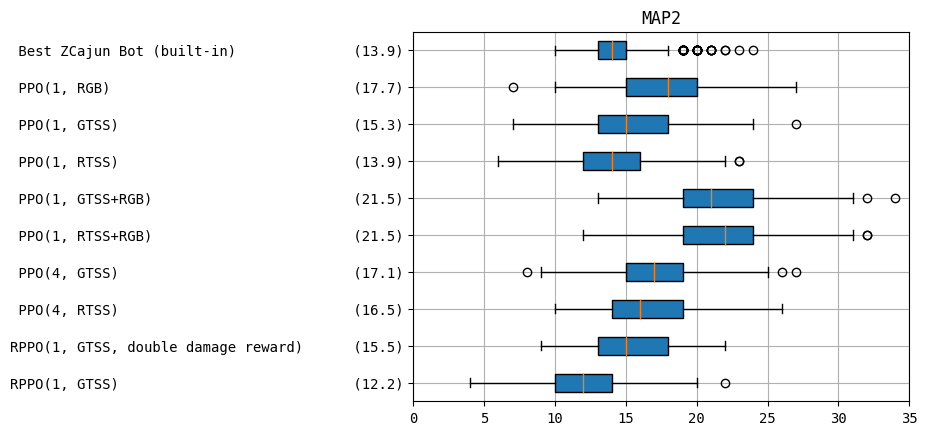}
    \caption{Evaluation Performance on Map 2.}
  \end{subfigure}
  \begin{subfigure}[b]{1.0\linewidth}
    \includegraphics[width=\linewidth]{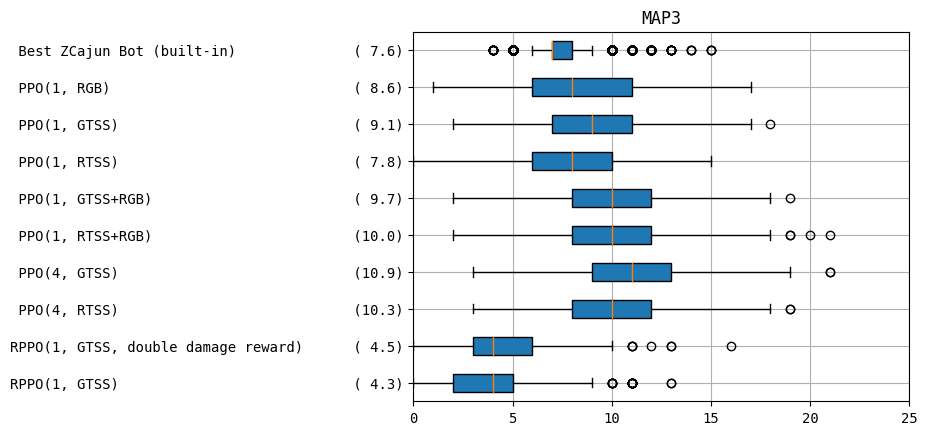}
    \caption{Evaluation Performance on Map 3.}
  \end{subfigure}
  \caption{Box-and-whisker plots for evaluation performance of RL agents on each map, with mean performance of agents enclosed in parentheses. \emph{\textbf{RTSS} refers to using our SS model for real-time semantic segmentation, \textbf{GTSS} refers to using perfect semantic segmentation (ground truth) from ViZDoom.}}
  \label{fig:eval-box}
\end{figure}

We conducted a hyperparameter search for learning rates across all PPO-based agents to identify the best-performing agent for each input representation. The evaluation results\footnote{All results are available on our GitHub repository: \href{https://github.com/Trenza1ore/SegDoom}{https://github.com/Trenza1ore/SegDoom}.} for these best agents are presented as box-and-whisker plots \cite{boxplot} in figure \ref{fig:eval-box}.

After 4 million training steps, all PPO-based agents significantly outperform our non-neural baseline: the ZCajun bots built into ViZDoom. These results are derived from 400 game-play sessions on each map for every PPO or RPPO-based agent. For comparison, the highest scores recorded in 4000 bot-only matches on each map were used to represent the non-neural baseline's performance.

\subsection{Performance on the Trained Map}
As a neural baseline, the RGB agent achieved an average frag\footnote{The scores in Doom deathmatches are known as "frags".} count of 15.2 on Map 1, exceeding the non-neural baseline's 9.8 by over 50\%. The RGB agent also had a median higher than ZCajun's non-outlier maximum and a 75th percentile equivalent to the maximum of ZCajun. This confirms the robustness of our neural baseline.

The SS-only agent yielded similar results with real-time and ground-truth semantic segmentation, with the real-time version's average frag count (15.0) only slightly lower than the 15.2 of ground-truth and RGB baseline, with a similar spread as showed by the same 25th and 75th percentile. SS(4) variants of the SS-only agent performed similarly with the additional temporal information, yielding an average frag count of 15.1 for ground-truth and 15.4 for real-time. All 4 SS-only agents yielded similar performance within the margin of error, with the same quartiles\footnote{25th/50th/75th percentiles} except for the real-time version of SS(4) which had slightly higher performance. This result suggests that for a task with moderate combat intensity and navigation complexity, the optimal policy learnt by SS-only agents would not benefit much from the additional temporal information.

RGB+SS outperformed every other agent by a significant amount with an average frag count of 19.3 and 19.2 for ground-truth and real-time respectively, with their 25th percentiles similar to the 75th percentile of every other PPO-based agent. The RGB+SS results confirm that augmenting the RGB input with additional semantic information greatly improves an RL agent's performance.

\subsection{Performance on the Trained Map with Textures Altered}
We hypothesised that changing the wall and floor textures in Map 1 as described in section \ref{changing-texture} would have some impact on RGB-based agents and believed that observing the changes in the performance of these agents would make an interesting side experiment. Screenshots of these two variants of Map 1 are displayed in appendix \ref{fig:map1-cmp}.

With only the wall textures altered, the RGB agent saw an 11\% decrease in average frag count while the RGB+SS agent only had its average decrease by 1\% with ground-truth SS and 5\% with real-time SS. The RGB+SS agent with ground-truth SS was not impacted much by the slight change in texture with its quartiles unchanged, as expected with an optimal policy that had perfect semantic information available in the input.

Changing the floor textures in addition to the wall textures created more impact despite the subtlety in the eyes of a human. The RGB agent had its average frag count unchanged from results in the previous Map 1 variant but the 75th percentile is decreased and despite sharing the same maximum, the maximum frag count in this variant of Map 1 became an outlier. The SS+RGB agent had its ground-truth performance decreased to close-to-real-time performance, which might indicate that the RGB part in the SS+RGB input helped to localise items on the ground.

More experiments with alternative textures would need to be carried out before any valid conclusion can be made but the observations from this side experiment do show that RGB+SS has better generalisation performance than the RGB baseline when the textures of the environments are changed by a small degree.

\subsection{Performance on Unseen Map with High Combat Intensity}
Map 2 is an open map with high combat intensity, where all players would receive a shotgun upon respawn instead of having to locate and pick up as in all the other maps. This high combat intensity is shown in the average frag counts in this map: all agents saw an increase in average frag count compared to map 1 except for the SS-only agent that used a real-time SS input, which yielded an average frag count of 13.9, the same as the non neural baseline.

Both SS-only agents and their SS(4) variants saw a decrease in average frag count when moving from ground-truth to real-time for semantic segmentation. However, while SS-only saw a 9\% decrease in average frag count and a decrease in every measurement (minimum, maximum and quartiles), the SS(4) variant only saw a small 3\% decrease in average frag count and the real-time version of SS(4) still have 75th percentile and maximum close to its ground-truth version. SS(4) variants with an average frag count of 16.5 for real-time and 17.1 for ground-truth are still comparable with the RGB neural baseline's 17.7, unlike the SS-only agents.

The neural baseline yielded a 17.7 average frag count in Map 2, with a 25th percentile equal to the 75th percentile of the non-neural baseline. The RGB+SS agents continue to outperform every other agent, with an average frag count of 21.5 for both real-time and ground-truth semantic segmentation and medians higher than the 75th percentile of all other agents.

Results from Map 2 indicate a clear advantage of having RGB input in scenarios with high combat intensity and no need for weapon acquirement (note that this also includes RGB+SS, which still outperformed every other agent by a significant amount). However, the temporal information presented in SS(4) input did help to make it comparable with the RGB baseline.

\subsection{Performance on Unseen Map with High Navigation Complexity}
Map 3 is a hard-to-navigate map with the lowest combat intensity and two hard-to-access shotguns, located at the small chambers at the upper right and lower left corners. Later heatmap analysis would reveal that agents tend to get trapped in this map a lot. The map also contained structures that were not present in Map 1, such as an unpassable see-through window, where the real-time semantic segmentation model would suffer.

The real-time SS-only agent continued to perform similarly to the non-neural baseline, with an average frag count of 7.8 compared to 7.6 of ZCajun. However, this agent has the same 25th percentile and median as the neural baseline of RGB, which had a slightly higher average frag count of 8.6. The ground-truth SS-only agent performed much better at the real-time version, yielding an average frag count of 9.1, indicating that the low quality in real-time semantic segmentation did hurt the performance of SS-only agents that only take a single predicted SS mask as input.

The RGB+SS agents performed similarly to the real-time version of SS(4), with the same values for quartiles and similar overall spread. The ground-truth version of SS(4) yielded the best performance, with an average frag count of 10.9 and the highest quartiles.

Results from Map 3 showed that the additional temporal information in SS(4) helped with navigation in complex scenes and that frame-stacking would increase the robustness of input representations that use SS masks only.

\section{Real-time Semantic Segmentation Performance}
Our semantic segmentation (SS) model employed to perform real-time SS for all experiments used ResNet-101, a 101-layer ResNet model as the encoder backbone to DeepLabV3. An Adam optimiser \cite{adam} was used for training, with an initial learning rate of 1e-4, scheduled to drop to 1e-5 after 70K steps. The model was trained for 20 epochs, yielding a final mIoU score of 0.846 on Map 1. \emph{The real-time SS model added a not negligible but still acceptable overhead to our RL agents, cutting the inference speed in half if a new instance is created for every vectorised environment.}

This result was lower than the reported 0.982 from a previous work \cite{PreviousSS} that used a slightly more complex model and higher input resolution (320$\times$240 instead of 256$\times$144), but was measured on real game-play sessions rather than on a validation subset of the same dataset used for training. As demonstrated in the previous section, our SS model proved to be sufficient for not introducing significant performance degradation on RL agents trained with perfect SS results at least on the map it was trained in.

\begin{table}[h!]
\centering
\resizebox{\columnwidth}{!}{
\begin{tabular}{|c|c|c|c|c|c|c|c|c|c|c|c|c|c|c|}
\hline
Map & mIoU & Floor. &   Wall & ItemF. & Telep. & Bulle. &  Blood &   Clip & Shell. & Shotg. & Medik. & DeadD. & DoomP. &   Self \\
\hline
1  &      0.846 &      0.994 &      0.979 &      0.657 &      0.703 &      0.326 &      0.380 &      0.294 &      0.667 &      0.552 &      0.676 &      0.803 &      0.753 &      0.957 \\
1w &      0.777 &      0.983 &      0.944 &      0.622 &      0.526 &      0.155 &      0.388 &      0.274 &      0.706 &      0.560 &      0.654 &      0.770 &      0.622 &      0.955 \\
1wf &      0.717 &      0.982 &      0.945 &      0.635 &      0.341 &      0.085 &      0.370 &      0.241 &      0.673 &      0.536 &      0.612 &      0.779 &      0.607 &      0.954 \\
2 &      0.693 &      0.982 &      0.970 &      0.558 &      0.509 &      0.112 &      0.236 &      NaN &      0.445 &      0.160 &      0.416 &      0.720 &      0.666 &      0.967 \\
3  &      0.683 &      0.981 &      0.908 &      0.510 &      0.290 &      0.032 &      0.320 &      0.117 &      NaN &      0.260 &      0.583 &      0.550 &      0.662 &      0.952 \\
\hline
\end{tabular}
}
\caption{IoU and mIoU measurement of the SS model. \emph{\textbf{1w} refers to the Map 1 variant with the wall textures altered and \textbf{1wf} means that both wall and floor textures are altered.}}
\label{tab:iou-cmp}
\end{table}

Table \ref{tab:iou-cmp} showed the IoU results of our SS model measured in 20 game-play episodes of our RGB+SS model on each map. The number of samples collected from each map varied from 95K to 100K due to variations in episode length (no observation can be made while the agent respawns after death). It can be observed that changing the floor texture by even a small amount degraded SS performance by a large amount for rare semantic classes like TeleportFog or BulletPuff, although the Floor/Ceiling class itself was barely affected. The poor identification accuracy of shotguns and dead doom players in Map 3 also contributed to the poor performance of real-time SS-only agents, leading to low weapon power and wasting of ammo on dead enemies respectively.

\section{Effectiveness of SS-only: Memory Usage Evaluation}
\begin{table}[h!]
\centering
\begin{tabular}{|c|c|c|c|c|c|}
\hline
Map & RGB & RGB (RLE) & SS & SS (RLE) & SS (RLE-EX)\\
\hline
1 (GT) & 11GB & 17GB (156.8\%) & 3.6GB (33.3\%) & 151MB (1.4\%) & 126MB (1.2\%) \\
1 (RT) & 11GB & 17GB (156.6\%) & 3.6GB (33.3\%) & 135MB (1.3\%) & 113MB (1.0\%) \\
2 (GT) & 11GB & 16GB (150.8\%) & 3.5GB (33.3\%) & 172MB (1.6\%) & 143MB (1.4\%) \\
2 (RT) & 11GB & 16GB (150.6\%) & 3.5GB (33.3\%) & 155MB (1.5\%) & 129MB (1.2\%) \\
3 (GT) & 11GB & 16GB (141.6\%) & 3.7GB (33.3\%) & 137MB (1.2\%) & 114MB (1.0\%) \\
3 (RT) & 11GB & 16GB (140.3\%) & 3.7GB (33.3\%) & 145MB (1.3\%) & 121MB (1.1\%) \\
\hline
\end{tabular}
\caption{Comparison of memory consumption for storing 20 episodes of game-play (without frame-skipping) for our best SS+RGB agent. This is equivalent to storing 80 episodes of game-play with a frame-skip value of 4. \emph{\textbf{RLE-EX} refers to utilising bitmasking to store the RLE elements in 4 bits, which adds a small overhead for decompression.}}
\label{tab:mem-cmp}
\end{table}

For our size comparison, a fully-vectorised implementation \cite{numpy-rle} of the run-length encoding (RLE) \cite{run-length-encoding} algorithm was used, storing the elements, length, starting position of runs in triplets of NumPy \cite{numpy} arrays. The encoding runs at more than 400 fps on our hardware\footnote{64GB DDR4 RAM, i9-12900HX CPU, RTX 3080Ti Mobile GPU with 16GB GDDR6 VRAM}, introducing an essentially negligible overhead. Decoding can happen directly on the GPU as filling operations, this overhead is also negligible as this replaces two operations: the casting to FP32\footnote{Data type for 32-bit floating-point numbers.} and the movement of data between CPU \& GPU memory. Example code for the compression and decompression with this variant of RLE can be found in appendix \ref{rle-code}. Reference implementation is available on \href{https://github.com/Trenza1ore/sb3-extra-buffers}{GitHub}.

\chapter{Conclusions}
In this project, we explored the feasibility of enhancing reinforcement learning (RL) in 3D environments through the use of semantic segmentation. Two novel representations were proposed and evaluated in three deathmatch scenarios of ViZDoom, each representing a different type of situation. The model adopted for semantic segmentation was a DeepLabV3 model with ResNet-101 backbone, which yielded an mIoU of 0.846 on the trained map and 0.693 \& 0.683 for the two unseen maps. The performance of our SS model was not exceptional but this model struck a good balance between the quality of semantic segmentation and inference speed overhead when applied to RL agents.

The \textbf{SS-only} input representation, as our solution to the high memory consumption problem in RL, yielded acceptable performance while taking only 1/3 of memory compared to the RGB baseline. Its performance was comparable to the RGB baseline model on the trained map while performing slightly worse in unseen maps where the semantic segmentation quality dropped. Results from our experiments implied that the RGB input is crucial to achieving high performance in scenarios with high combat intensity, an area where SS-only suffers. The attempt at applying a lossless compression technique (run-length encoding) to SS-only was a major success, further reducing memory consumption to less than 2\% of RGB baseline while introducing minimal compression overhead. The memory-saving nature of SS-only allowed us to employ the frame-stacking technique and stack multiple observations together to provide temporal information to an agent, allowing it to navigate complex maps better than other agents.

Our \textbf{RGB+SS} input representation combined the advantages of SS-only and RGB input, achieving good performance in combat-intense situations while also using the semantic information to guide its action, resulting in it outperforming every other input representation without frame-stacking. However, the frame-stacking variant of SS-only can still outperform it by a small amount in scenarios of high navigation complexity due to the presence of temporal information.

The tools and techniques developed for this project can potentially prove to be beneficial to other projects involving RL agents in 3D environments, such as the density-based heatmapping approach for visualisation and vectorised run-length encoding compression for semantic segmentation masks.

\section{Future Works}
Many directions can be expanded upon for future works. One of them is the usage of SS-only representation in off-policy RL models that made use of experience replay, where one of the major difficulties when working with limited memory was that the model would not be able to learn to perform high-reward actions before the replay memory is filled with useless low-reward state-action pairs and the model could never learn the task. With the low memory usage of compressed SS-only representation, more experiences can be stored in the replay memory, giving the model more time to learn.

Another area to explore is using RPPO agents with shared LSTMs, recurrent models typically perform better than non-recurrent ones in POMDP situations, as proven by the previous state-of-the-art in ViZDoom. However, the hyperparameter tuning can still be very demanding.

Finally, it would also be interesting to conduct more experiments on the semantic segmentation model itself, our SS model was less than ideal due to the time constraint of this project and for simplicity did not experiment on any down-sampling technique.

\bibliographystyle{plain}
\bibliography{mybibfile}

\appendix

\chapter{Additional Code}
This part contains the additional source code and pseudocode that couldn't fit into the main report within the page limitations.
\newpage
\label{rle-code}
\section{Vectorised RLE Implementation in NumPy \cite{numpy}}
This code snippet is taken from the original post \cite{numpy-rle} with longer comments trimmed.
\begin{lstlisting}
import numpy as np

def rle(inarray):
        """ run length encoding. Partial credit to R rle function. 
            Multi datatype arrays catered for including non Numpy
            returns: tuple (runlengths, startpositions, values) """
        ia = np.asarray(inarray)                # force numpy
        n = len(ia)
        if n == 0: 
            return (None, None, None)
        else:
            y = ia[1:] != ia[:-1]               # pairwise unequal
            i = np.append(np.where(y), n - 1)   # include last element pos
            z = np.diff(np.append(-1, i))       # run lengths
            p = np.cumsum(np.append(0, z))[:-1] # positions
            return(z, p, ia[i])
\end{lstlisting}

\section{Naive Implementation for Reconstruction from RLE}
This code snippet is an example of how RLE-compressed arrays can be reconstructed.
\begin{lstlisting}
import numpy as np

length, pos, elements = rle(arr)                # 1D example
arr_reconstructed = np.empty_like(arr)          # can be a tensor on GPU

for start_pos, end_pos, elem in zip(pos, pos + length, elements):
    arr_reconstructed[start_pos:end_pos] = elem # simple filling operation

print(np.all(arr == arr_reconstructed))         # > True
\end{lstlisting}

\newpage
\section{Pseudocode for Vectorised Run-Length Encoding}
\begin{algorithm}[h!]
\caption{Vectorised RLE by Checking Each Possible Symbol Individually}
\label{alg:vec_rle}
\begin{algorithmic}[1]
\Require Input array $A$ of length $n$, and set of possible symbols $S=\{S_1, S_2, \dots, S_k\}$
\Ensure Integer $i$; Arrays $\text{symbols}$, $\text{start\_positions}$, and $\text{lengths}$ of length $n$

\State $\text{symbols} \gets \text{empty array}[0:n-1]$
\State $\text{start\_positions} \gets \text{empty array}[0:n-1]$
\State $\text{lengths} \gets \text{empty array}[0:n-1]$
\State $i \gets 0$

\For{each symbol $S_j$ in $S$} \Comment{Vectorised loop over all possible symbols}
    \State $\text{mask} \gets (A = S_j)$ \Comment{Vectorised comparison: mask of where $A$ equals to $S_j$}
    \State $\text{run\_starts} \gets \text{find\_start\_positions}(\text{mask})$ \Comment{Find start positions of runs (async)}
    \State $\text{run\_lengths} \gets \text{find\_run\_lengths}(\text{mask})$ \Comment{Find the lengths of runs (async)}
    
    \For{each start position $p$ in $\text{run\_starts}$}
    \Comment{As results return asynchronously}
        \State Wait for corresponding length $l$ from $\text{run\_lengths}$ to be returned
        \State Acquire $\textbf{thread lock}$
        \State $\text{symbols}[i] \gets S_j$
        \State $\text{start\_positions}[i] \gets p$
        \State $\text{lengths}[i] \gets l$
        \State $i \gets i + 1$
        \State Release $\textbf{thread lock}$
    \EndFor
\EndFor

\State Truncate $\text{symbols}$, $\text{start\_positions}$, and $\text{lengths}$ to a length of $i$

\end{algorithmic}
\end{algorithm}

\chapter{Additional Images}
This part contains the additional images that couldn't fit into the main report within the page limitations.
\newpage

\section{Alternative Textures for Map 1}
\begin{figure}[b!]
  \centering
  \begin{subfigure}[b]{0.75\linewidth}
    \includegraphics[width=\linewidth]{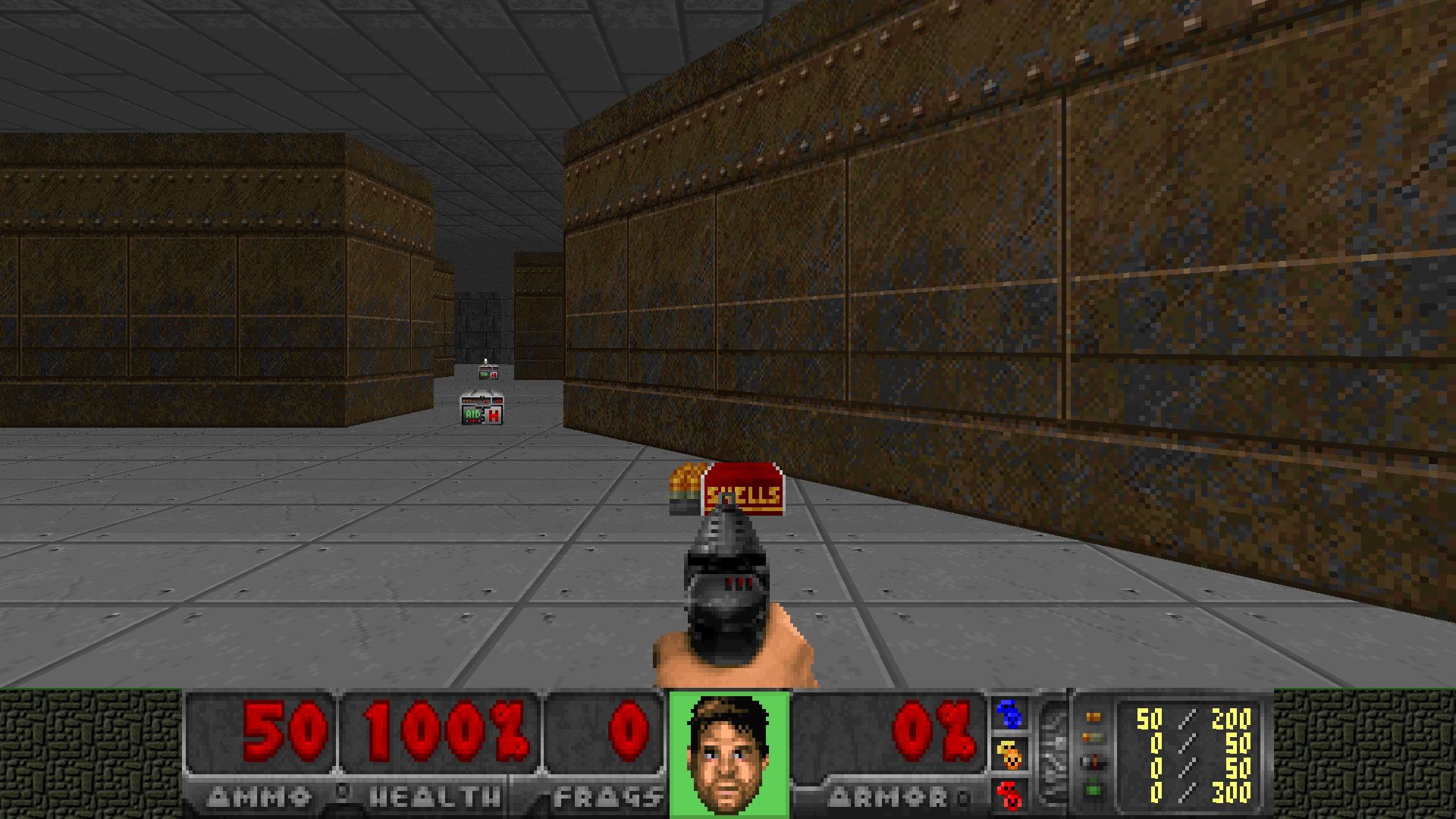}
    \caption{Original.}
  \end{subfigure}
  \begin{subfigure}[b]{0.75\linewidth}
    \includegraphics[width=\linewidth]{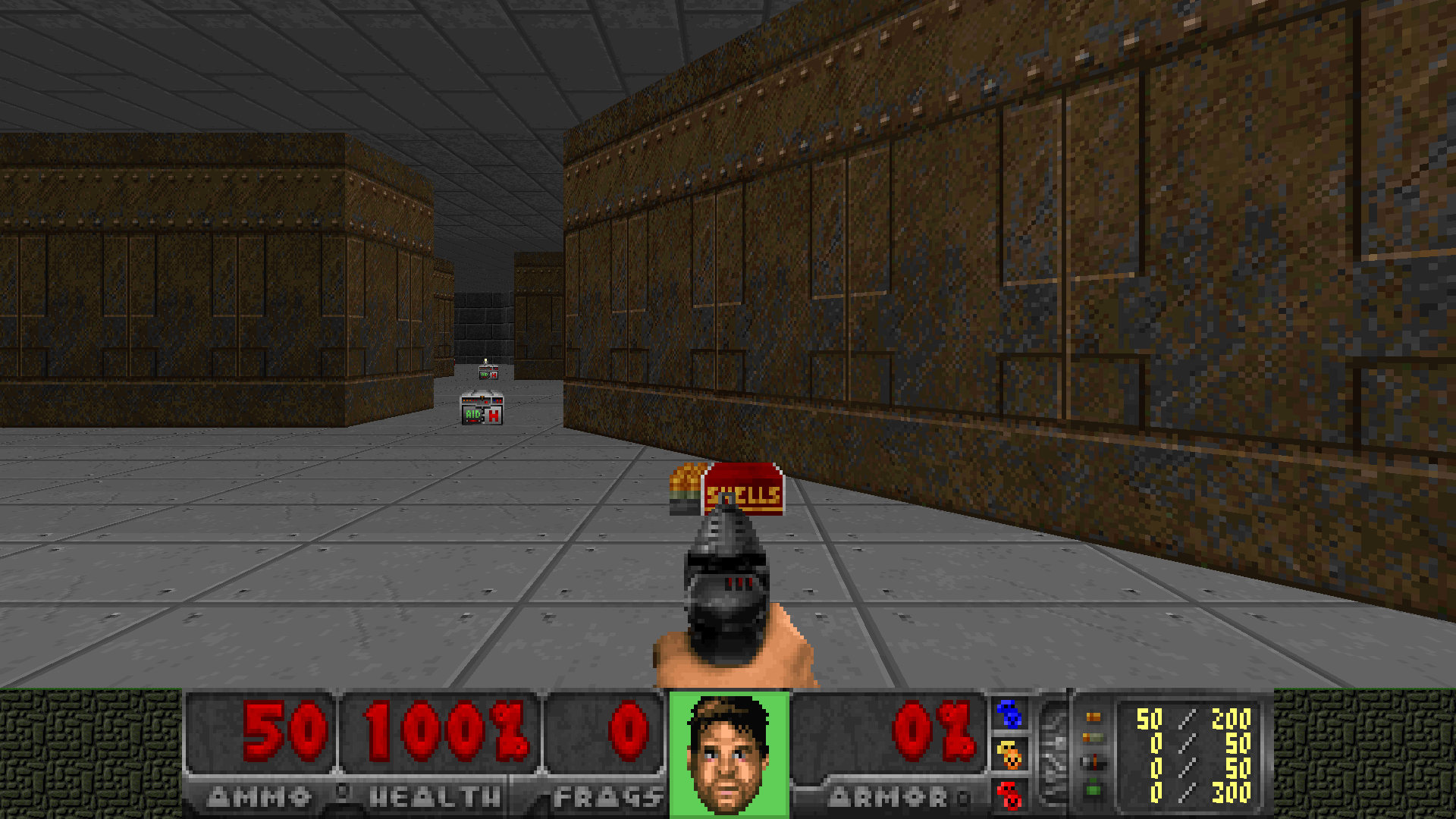}
    \caption{Wall textures altered.}
  \end{subfigure}
  \begin{subfigure}[b]{0.75\linewidth}
    \includegraphics[width=\linewidth]{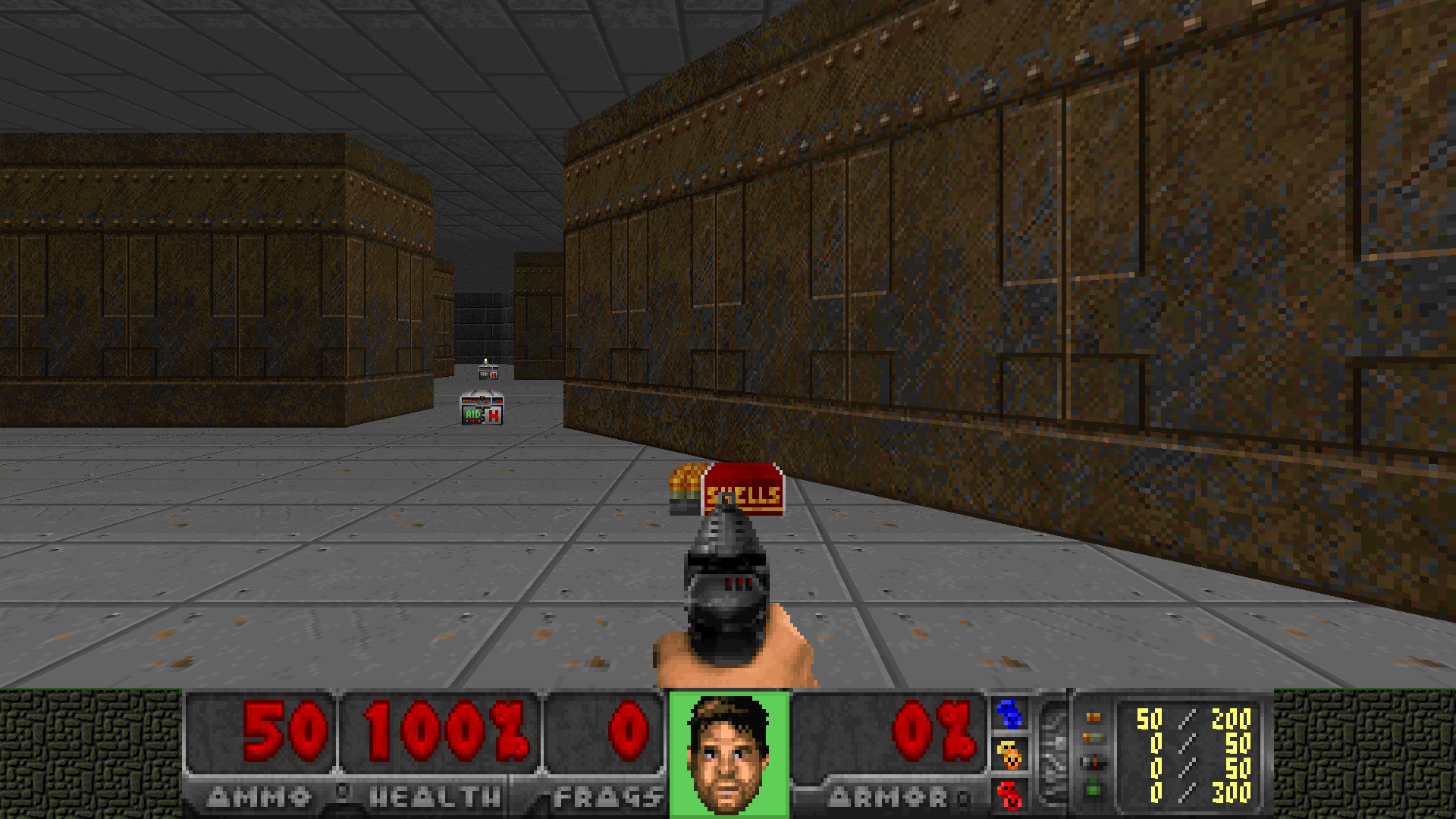}
    \caption{Wall and floor textures altered.}
  \end{subfigure}
  \caption{Alternative textures for Map 1.}
  \label{fig:map1-cmp}
\end{figure}

\chapter{Additional Tables}
This part contains the additional tables that couldn't fit into the main report within the page limitations.
\section{Weapon Usage}
\begin{itemize}
    \item It can be observed that doubling the reward for killing and damaging enemies encouraged the RPPO agents to pick up shotguns more often.
    \item More usage of the shotgun, in general, meant better performance as it is a much stronger weapon than the default pistol.
    \item The altered textures in variants of Map 1 did not affect the pick-up of shotguns by any meaningful amount.
    \item The low shotgun pick-up rate in Map 3 is likely the reason behind RGB+SS's lower performance, as its shotgun usage was above 82\% in the other maps and it likely had learnt to use shotgun efficiently.
\end{itemize}

\begin{table}[h]
\centering
\begin{tabular}{|l|l|c|c|}
\hline
RL Agent & Map & \textbf{Shotgun\%} & Pistol\%\\
\hline
\multirow{4}{*}{PPO(1, RGB)} & map1 & 79 & 21 \\
& map1 (alt. wall) & 77 & 23 \\
& map1 (alt. wall \& floor) & 79 & 21 \\
& map3 & 74 & 26 \\
\hline
\multirow{8}{*}{PPO(1, RGB+SS)} & map1 (gt) & 83 & 17 \\
& map1 (rt)& 83 & 17 \\
& map1 (rt, alt. wall) & 83 & 17 \\
& map1 (gt, alt. wall) & 83 & 17 \\
& map1 (rt, alt. wall \& floor) & 82 & 18 \\
& map1 (gt, alt. wall \& floor) & 82 & 18 \\
& map3 (gt) & 62 & 38 \\
& map3 (rt) & 62 & 38 \\
\hline
\multirow{4}{*}{PPO(1, SS)} & map1 (gt) & 78 & 22 \\
& map1 (rt) & 76 & 24 \\
& map3 (gt) & 72 & 28 \\
& map3 (rt) & 72 & 28 \\
\hline
\multirow{4}{*}{PPO(4, SS)} & map1 (gt) & 71 & 29 \\
& map1 (rt) & 72 & 28 \\
& map3 (gt) & 70 & 30 \\
& map3 (rt) & 69 & 31 \\
\hline
\multirow{2}{*}{RPPO(1, SS)} & map1 (gt) & 36 & 64 \\
& map3 (gt) & 56 & 44 \\
\hline
\multirow{2}{*}{RPPO(1, SS, 2$\times$DR)} & map1 (gt) & 66 & 34 \\
& map3 (gt) & 59 & 41 \\
\hline
\end{tabular}
\caption{Weapon usage data of different agents.
\emph{\textbf{gt} refers to using ground truth for semantic segmentation as opposed to \textbf{rt} which uses prediction from the SS model; 
\textbf{2$\times$DR} refers to doubling the reward for kills and shaping reward for damaging enemies during training; 
\textbf{alt.} refers to the alternation of textures in the original map.}
}
\label{tab:wpn-usage}
\end{table}

\chapter{Heatmaps}
The heatmaps allow us to visualise the difference in movement patterns of different agents in each map. Each agent has its unique characteristics but also shares many similarities. For example, all heatmaps of Map 1 showed clear traces connecting spawn points to weapons (shotguns), indicating that all well-performing agents learnt to pick up the shotgun as soon as it respawns. Some interesting patterns can be observed in the chambers of Map 3 that contained the shotgun (where a frame-skip of 4 might be too high and the agent could not turn around before hitting the wall) and the L-shape/C-shape structure where the see-through window structure is located (the agents did not train with this kind of structure and got stuck there a lot).
\begin{figure}[h]
  \centering
  \begin{subfigure}[b]{0.45\linewidth}
    \includegraphics[width=\linewidth]{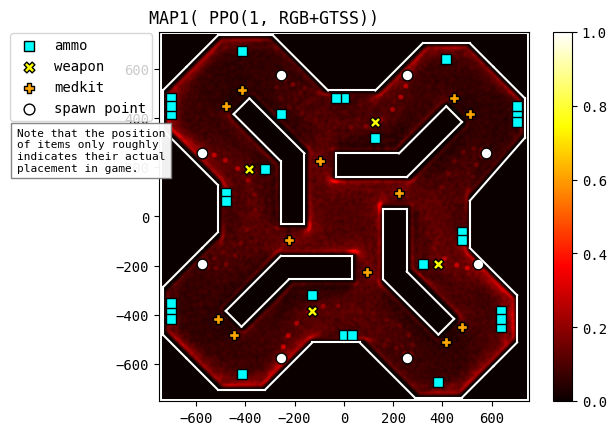}
    \caption{GT RGB+SS on Map 1.}
  \end{subfigure}
  \begin{subfigure}[b]{0.45\linewidth}
    \includegraphics[width=\linewidth]{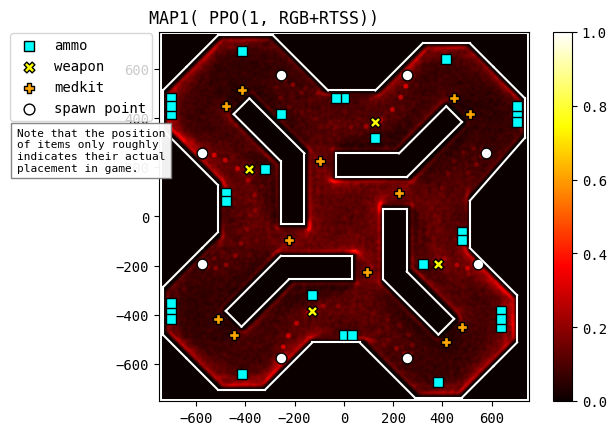}
    \caption{RT RGB+SS on Map 1.}
  \end{subfigure}
  \begin{subfigure}[b]{0.45\linewidth}
    \includegraphics[width=\linewidth]{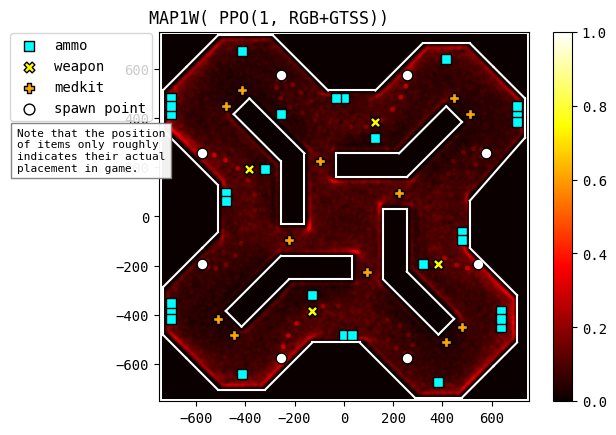}
    \caption{GT RGB+SS on Map 1 (alt. wall).}
  \end{subfigure}
  \begin{subfigure}[b]{0.45\linewidth}
    \includegraphics[width=\linewidth]{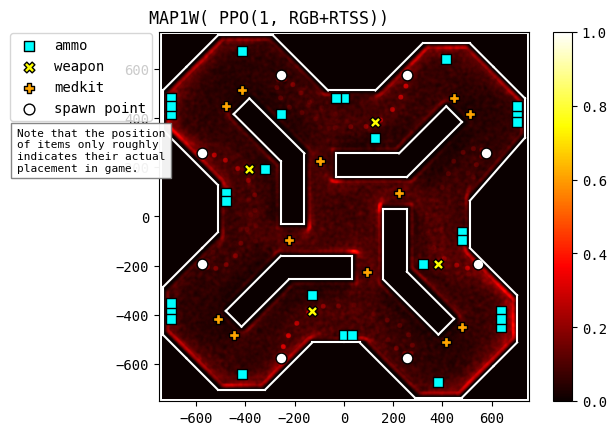}
    \caption{RT RGB+SS on Map 1 (alt. wall).}
  \end{subfigure}
  \begin{subfigure}[b]{0.45\linewidth}
    \includegraphics[width=\linewidth]{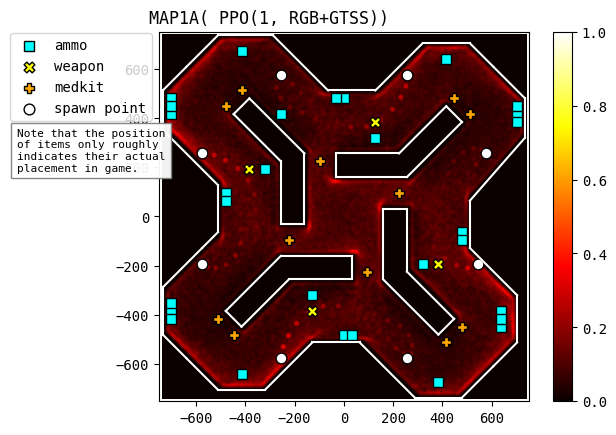}
    \caption{GT RGB+SS on Map 1 (alt. wall \& floor).}
  \end{subfigure}
  \begin{subfigure}[b]{0.45\linewidth}
    \includegraphics[width=\linewidth]{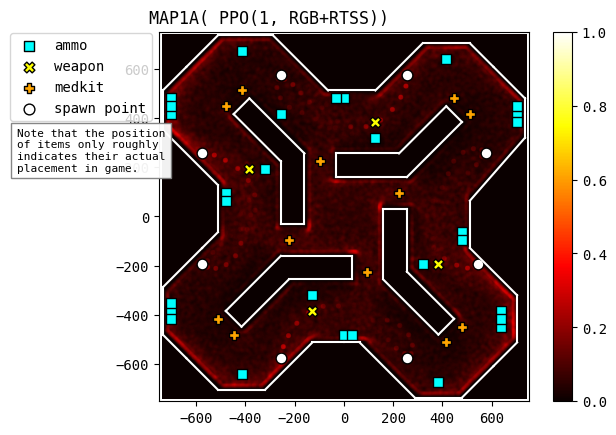}
    \caption{RT RGB+SS on Map 1 (alt. wall \& floor).}
  \end{subfigure}
  \begin{subfigure}[b]{0.45\linewidth}
    \includegraphics[width=\linewidth]{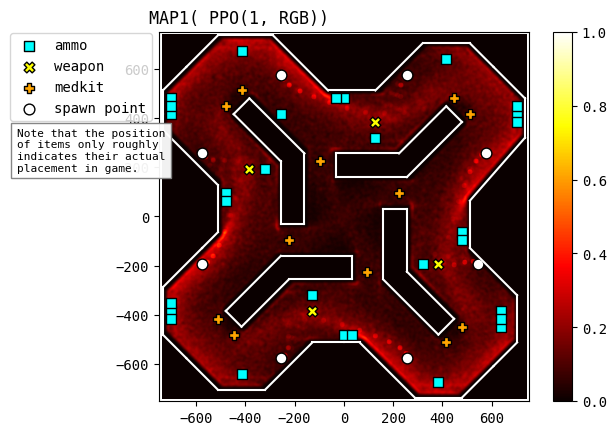}
    \caption{RGB on Map 1.}
  \end{subfigure}
  \begin{subfigure}[b]{0.45\linewidth}
    \includegraphics[width=\linewidth]{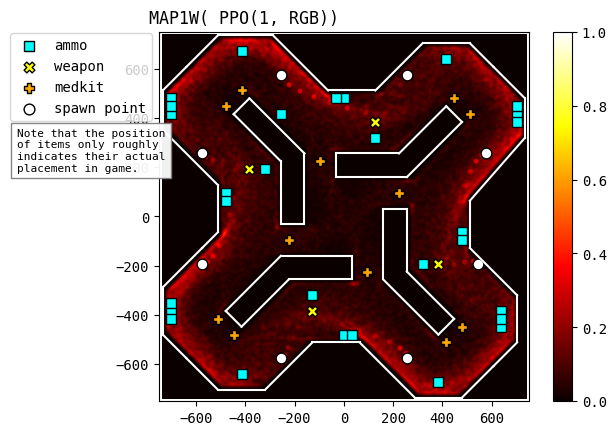}
    \caption{RGB on Map 1 (alt. wall).}
  \end{subfigure}
  \begin{subfigure}[b]{0.45\linewidth}
    \includegraphics[width=\linewidth]{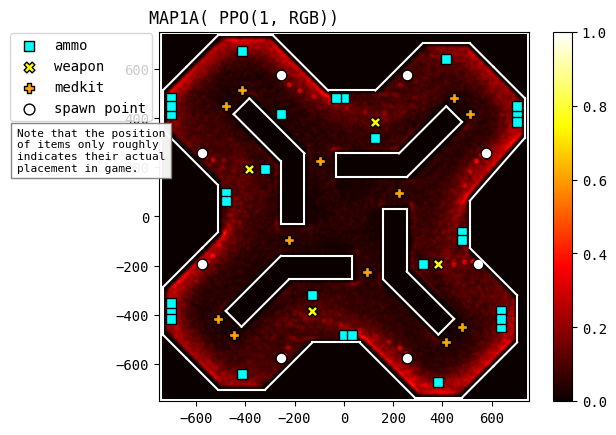}
    \caption{RGB on Map 1 (alt. wall \& floor).}
  \end{subfigure}
  \label{fig:heatmap1}
\end{figure}

\begin{figure}[h]
  \centering
  \begin{subfigure}[b]{0.45\linewidth}
    \includegraphics[width=\linewidth]{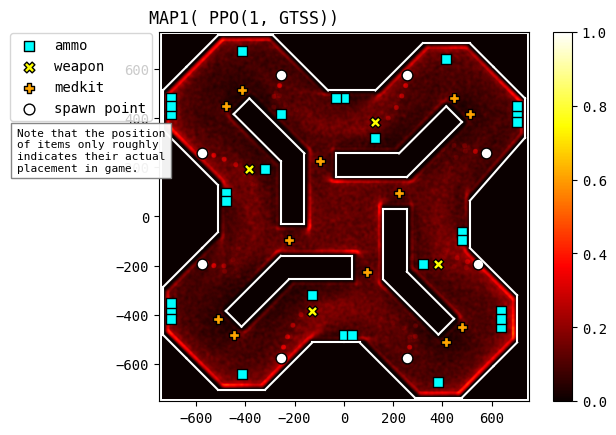}
    \caption{GT SS-only on Map 1.}
  \end{subfigure}
  \begin{subfigure}[b]{0.45\linewidth}
    \includegraphics[width=\linewidth]{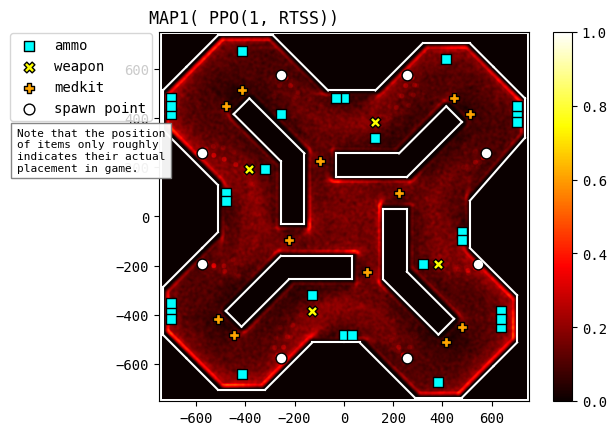}
    \caption{RT SS-only on Map 1.}
  \end{subfigure}
  \begin{subfigure}[b]{0.45\linewidth}
    \includegraphics[width=\linewidth]{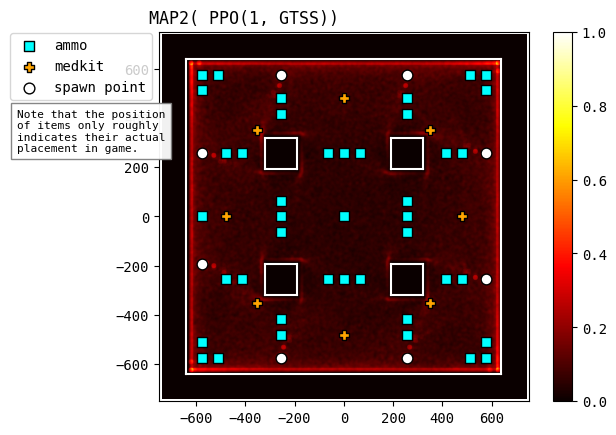}
    \caption{GT SS-only on Map 2.}
  \end{subfigure}
  \begin{subfigure}[b]{0.45\linewidth}
    \includegraphics[width=\linewidth]{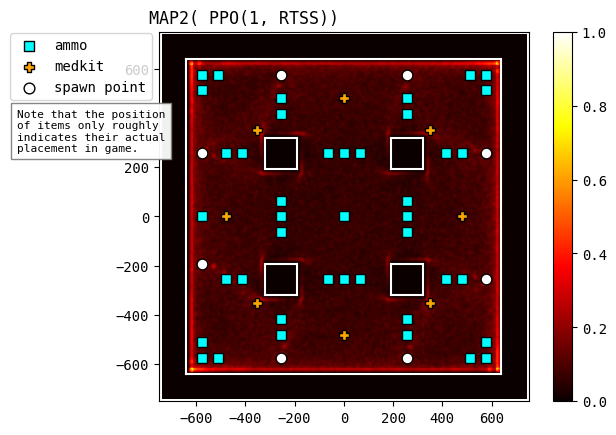}
    \caption{RT SS-only on Map 2.}
  \end{subfigure}
  \begin{subfigure}[b]{0.45\linewidth}
    \includegraphics[width=\linewidth]{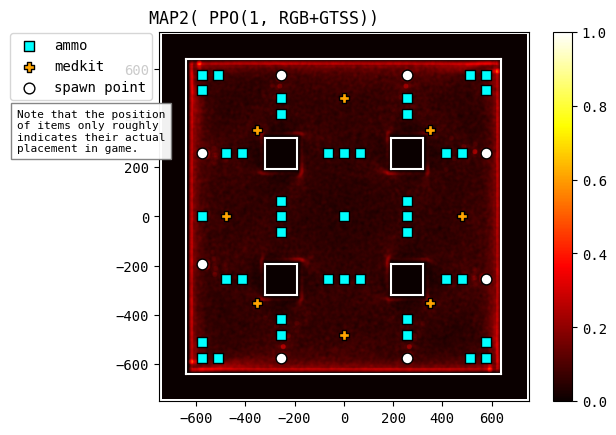}
    \caption{GT RGB+SS on Map 2.}
  \end{subfigure}
  \begin{subfigure}[b]{0.45\linewidth}
    \includegraphics[width=\linewidth]{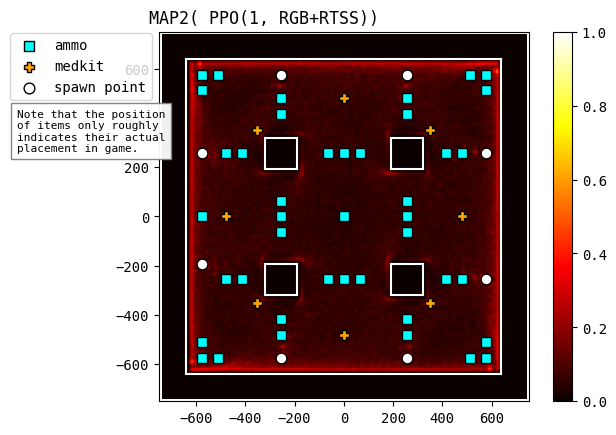}
    \caption{RT RGB+SS on Map 2.}
  \end{subfigure}
  \begin{subfigure}[b]{0.45\linewidth}
    \includegraphics[width=\linewidth]{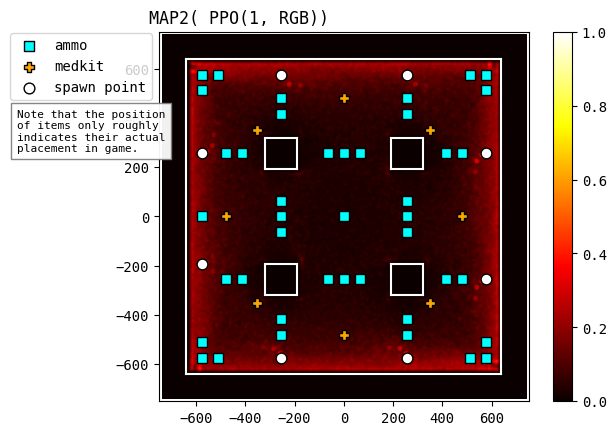}
    \caption{RGB on Map 2.}
  \end{subfigure}
  \label{fig:heatmap2}
\end{figure}

\begin{figure}[h]
  \centering
  \begin{subfigure}[b]{0.45\linewidth}
    \includegraphics[width=\linewidth]{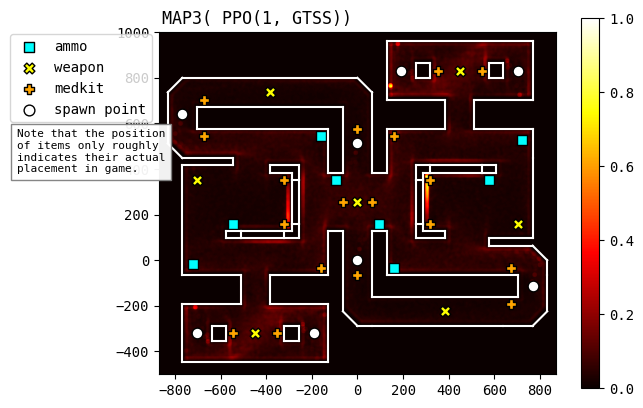}
    \caption{GT SS-only on Map 3.}
  \end{subfigure}
  \begin{subfigure}[b]{0.45\linewidth}
    \includegraphics[width=\linewidth]{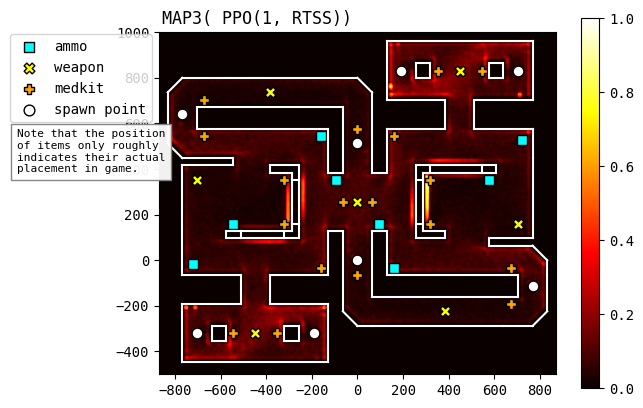}
    \caption{RT SS-only on Map 3.}
  \end{subfigure}
  \begin{subfigure}[b]{0.45\linewidth}
    \includegraphics[width=\linewidth]{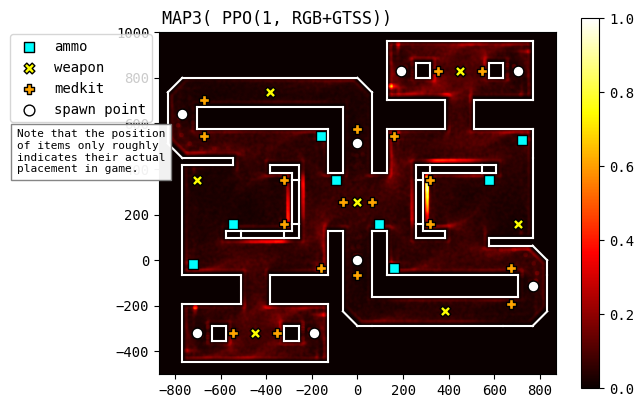}
    \caption{GT RGB+SS on Map 3.}
  \end{subfigure}
  \begin{subfigure}[b]{0.45\linewidth}
    \includegraphics[width=\linewidth]{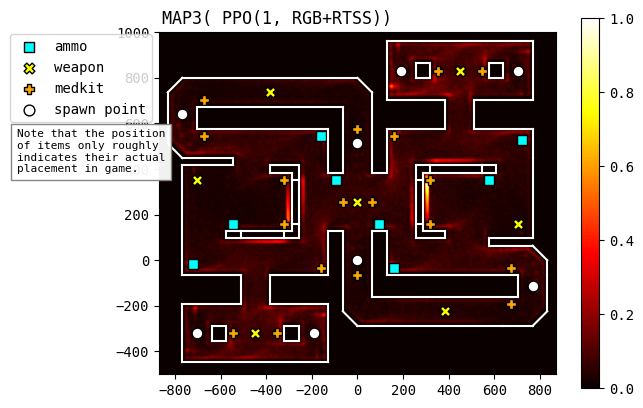}
    \caption{RT RGB+SS on Map 3.}
  \end{subfigure}
  \begin{subfigure}[b]{0.45\linewidth}
    \includegraphics[width=\linewidth]{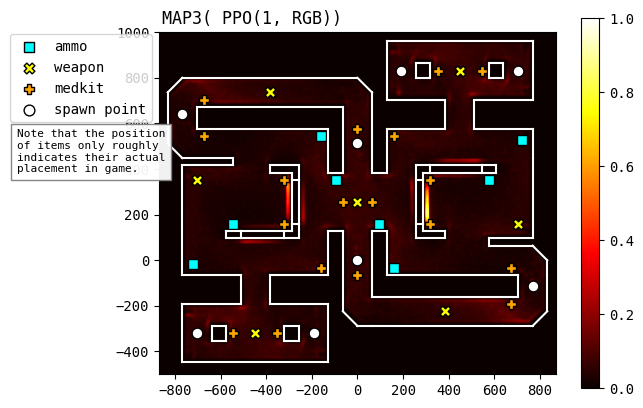}
    \caption{RGB on Map 3.}
  \end{subfigure}
  \label{fig:heatmap3}
\end{figure}

\end{document}